\documentclass[lettersize,journal]{IEEEtran}
\usepackage{amsmath,amsfonts}
\usepackage{algorithmic}
\usepackage{algorithm}
\usepackage{array}
\usepackage[caption=false,font=normalsize,labelfont=sf,textfont=sf]{subfig}
\usepackage{textcomp}
\usepackage{stfloats}
\usepackage{url}
\usepackage{verbatim}
\usepackage{graphicx}
\usepackage{cite}
\usepackage{xspace}         % special space based on context
\usepackage[inline]{enumitem}
\usepackage{pifont}
\usepackage{amsfonts,amsmath}
\usepackage{booktabs}
\usepackage{graphicx}
\usepackage{multirow}
\usepackage{adjustbox}
\usepackage[dvipsnames,table,xcdraw]{xcolor}
\usepackage{subcaption}

\newcommand{\name}{\textsc{D2DGN}\xspace}

\newcommand{\G}{\mathcal{G}\xspace}
\newcommand{\V}{\mathcal{V}\xspace}
\newcommand{\E}{\mathcal{E}\xspace}
\newcommand{\D}{\mathcal{D}\xspace}

\usepackage{tikz}

\begin{document}
\title{Distill to Delete: Unlearning in Graph Networks with Knowledge Distillation}
%\author{IEEE Publication Technology,~\IEEEmembership{Staff,~IEEE,}
        % <-this % stops a space

%\author{}
%\maketitle
\makeatletter
\newcommand{\printfnsymbol}[1]{%
  \textsuperscript{\@fnsymbol{#1}}%
}
\makeatother
\author{    \IEEEauthorblockN{Yash~Sinha\IEEEauthorrefmark{1}, Murari~Mandal\IEEEauthorrefmark{2}, Mohan~Kankanhalli\IEEEauthorrefmark{1}}\\
\IEEEauthorblockA{\IEEEauthorrefmark{1}School of Computing, National University of Singapore}\\
\IEEEauthorblockA{\IEEEauthorrefmark{2}RespAI Lab, KIIT Bhubaneswar, India}\\
yashsinha@comp.nus.edu.sg, murari.mandalfcs@kiit.ac.in, mohan@comp.nus.edu.sg}
%\author{Yash~Sinha, Murari~Mandal, Mohan~Kankanhalli}
%\thanks{\textsuperscript{$\ddagger$}\textit{Corresponding Author}. Work performed while at the School of Computing, National University of Singapore.}
%\thanks{Ayush~K~Tarun, Vikram~S~Chundawat are with Mavvex Labs, India, Faridabad 121001 (ayushtarun210@gmail.com; vikram2000b@gmail.com), Murari~Mandal is with the School of Computer Engineering, Kalinga Institute of Industrial Technology (KIIT) Bhubaneswar, India 751024 (Email: murari.mandalfcs@kiit.ac.in), Mohan~Kankanhalli is with the School of Computing, National University of Singapore (NUS), Singapore 117417 (Email: mohan@comp.nus.edu.sg)}
%}
\maketitle
\begin{abstract}
Graph unlearning has emerged as a pivotal method to delete information from an already trained graph neural network (GNN). One may delete nodes, a class of nodes, edges, or a class of edges. An unlearning method enables the GNN model to comply with data protection regulations (i.e., the right to be forgotten), adapt to evolving data distributions, and reduce the GPU-hours carbon footprint by avoiding repetitive retraining. Removing specific graph elements from graph data is challenging due to the inherent intricate relationships and neighborhood dependencies. Existing partitioning and aggregation-based methods have limitations due to their poor handling of local graph dependencies and additional overhead costs. 
% More recently, \textsc{GNNDelete} offered a model-agnostic approach that alleviates some of these issues. 
Our work takes a novel approach to address these challenges in graph unlearning through knowledge distillation, as it \textit{distills to delete} in GNN (\name). It is an efficient model-agnostic distillation framework where the complete graph knowledge is divided and marked for retention and deletion. It performs distillation with response-based soft targets and feature-based node embedding while minimizing KL divergence. The unlearned model effectively removes the influence of the deleted graph elements while preserving knowledge about the retained graph elements. \name surpasses the performance of existing methods when evaluated on various real-world graph datasets by up to $\mathbf{43.1\%}$ (AUC) in edge and node unlearning tasks. Other notable advantages include better efficiency, better performance in removing target elements, preservation of performance for the retained elements, and zero overhead costs. 
% Notably, our \name surpasses the state-of-the-art \textsc{GNNDelete} in AUC by $2.4\%$, improves membership inference ratio by $+1.3$, requires $10.2 \times 10^6$ fewer FLOPs per forward pass and up to $\mathbf{3.2}\times$ faster.
\end{abstract}
\section{Introduction} \label{sec:intro}
Graph data is ubiquitous and pervasive in today's data-driven world, capturing intricate relationships and dependencies among elements. Graphs represent the complex structures in various domains like social networks, biological systems, knowledge graphs, and recommendation systems~\cite{hu2020open}. Graph neural networks (GNNs) are powerful tools for analyzing and processing graph-based information. Through applications in healthcare, natural language processing, financial systems, and transportation~\cite{wu2020comprehensive},~\cite{zhou2020graph}, GNNs have showcased their potential to discover valuable insights and enable technological advancements. However, like any other machine learning (ML) algorithm, GNNs also encode information about the training data. The recently introduced data protection regulations like GDPR~\cite{voigt2017eu}, CCPA~\cite{goldman2020introduction}, and LGPD~\cite{hoffmann2022lgpd} grant the individual data holder \textit{the right to be forgotten}, allowing them to protect their private, sensitive, and identifying information. Machine unlearning algorithms facilitate such deletion of data upon request in an ML model~\cite{nguyen2022survey}. For graph data, graph unlearning has emerged as a pivotal method to delete information from an already trained GNN model. 
%, thus enabling GNNs to adapt to upcoming requirements for legal compliance under these laws.

In addition to legal compliance, graph unlearning has the potential to bring forth other advantages. It can enable the removal of biased, incomplete, or malicious data, thus promoting fairness and transparency. Moreover, it can facilitate the adaptation of models for evolving data distributions and changing trends. Most importantly, it can reduce the GPU-hours carbon footprint by avoiding repetitive retraining. However, designing an effective graph unlearning method presents several challenges that demand careful consideration, as the elements in a graph structure share inherent intricate relationships and neighborhood dependencies.\par

\textbf{Background.} The existing conventional unlearning methods~\cite{bourtoule2021machine} partition the graph dataset into multiple shards. This breaks the graph structure, which leads to increased costs and sub-optimal performance~\cite{chen2022graph,wang2023inductive}. Performance in each shard suffers from insufficient training data and poor data heterogeneity. Additional strategies of preserving graph structure and aggregating the performance of individual shards add high overhead costs, affecting the training and inference procedures significantly~\cite{koch2023no}. The cost increases with the increasing number of shards~\cite{ramezani2021learn}. 

%This way, the model retains the influence of deleted elements, resulting in an ineffective unlearning process. It may potentially lead to privacy breaches by revealing whether specific data points were present in its training dataset~\cite{olatunji2021membership}.

%This way, the model retains the influence of deleted elements, resulting in an ineffective unlearning process.

As the graph elements share similar properties and features, the unlearned model may still have retained elements that implicitly encode the deleted elements' features, potentially leading to privacy breaches~\cite{olatunji2021membership}. Trying to overcome these issues may inadvertently undermine the performance of the retained elements due to a solid dependency on the deleted local graph neighborhoods. Striking a delicate balance between graph structure, privacy preservation, and model performance has led to the design of various methods. Some methods tailor unlearning specifically for linear GNNs using techniques like ridge regression \cite{cong2022grapheditor} or projection \cite{cong2023efficiently}. Others approximate the learning process through influence functions \cite{wu2023gif} or theoretical guarantees \cite{chien2022certified}, \cite{chien2023efficient}, \cite{wu2023certified}. However, these approaches are often GNN-specific, demand prior knowledge of the training procedure, and scale poorly with processing large unlearning requests. While strategies like sub-graph sampling enhance scalability, they introduce additional costs to the training and inference procedures. Therefore, a model-agnostic graph unlearning method which has minimal overhead costs and is efficient in time as compared to the retrained, or \textit{gold} model is highly desired~\cite{nguyen2022survey}. More recently, \textsc{GNNDelete}~\cite{cheng2023gnndelete} presents a model-agnostic approach that alleviates some of these issues. However, it introduces an additional deletion layer, which incurs additional time and costs for model inference. 
% The supplementary material delves further into the details regarding the related work \cite{pan2023unlearning},~\cite{chundawat2022zero},~\cite{tarun2023deep},~\cite{zhu2023heterogeneous},~\cite{chundawat2023can},~\cite{dukler2023safe},~\cite{tarun2021fast}.

\textbf{Our Contribution.} Our work takes a novel approach to address the challenges mentioned above in graph unlearning through knowledge distillation, as in we \textit{distill to delete} in GNN, or in short, \name. To the best of our knowledge, this is the first attempt to unlearn GNN with distillation. This approach has three key aspects: distillation architecture, distillation measure, and knowledge type. 
\name's distillation architecture introduces two separators for distilling knowledge from the source model. The preserver, a pre-trained model, distills knowledge marked for retention. The destroyer, either a randomly initialized, untrained, or fully trained model, distills the knowledge marked for deletion. \name distills different knowledge types: response-based soft targets and feature-based node embeddings, using distillation measures of Kullback-Leibler Divergence loss and Mean Squared Error loss, respectively. We evaluate our \name on five real-world graph datasets with varied sizes and characteristics. In comparison with the \textsc{Gold} model, which is retrained from scratch using retained data as well as existing state-of-the-art methods, \name exhibits superior performance across various aspects:
\begin{enumerate}%[label=(\arabic*)]
  \item \textit{Consistency}: \name effectively deletes or unlearns the influence of elements marked for deletion. On the forget set, it achieves accuracy closest to the \textsc{Gold} model, as close as $\delta=0.5\%$. While the \textsc{Gold} model has an accuracy of $50.6\%$, \name achieves $50.1\%$.% The best accuracy achieved among the baselines is $50.0\%$.
  
  \item \textit{Integrity}: \name preserves the original performance of the model in the local vicinity of the unlearned element. It achieves accuracy closest to the \textsc{Gold} model, as close as $\delta=0.6\%$. While the \textsc{Gold} model has an accuracy of $96.4\%$, \name achieves $95.8\%$. The previous state-of-the-art \textsc{GNNDelete} achieves $93.4\%$, while other baselines achieve up to $71\%$.

  \item \textit{Membership Privacy}: \name outperforms all baselines on membership inference (MI) attack \cite{olatunji2021membership}, improving the MI ratio by $+1.3$, highlighting its effectiveness in preventing leakage of deleted data.

  %\item Efficiency: \name is efficient in time and space. %Particularly, \name requires $6.291 \times 10^6$ fewer FLOPs per forward pass, making it lightweight.

  \item \textit{Unlearning Cost \& Inference Cost}: \name has zero partitioning or aggregation overhead costs. It does not require prior knowledge of the training procedure or optimization techniques. Notably, \name requires $10.2 \times 10^6$ fewer FLOPs than \textsc{GNNDelete} per forward pass, making it lightweight and up to $\mathbf{3.2}\times$ faster.

 %\item Generality: \name \textcolor{blue}{applies to various tasks, covering various aspects of model usage, such as unlearning nodes, node classes, edges, and edge classes. Add results.}
\end{enumerate}

\section{Related Work}
\label{sec:related}
% \subsection{Machine Unlearning}
% Recent methods for unlearning (Bourtoule et al., 2021; Wu et al., 2020; Liu et al., 2022b; Cao \& Yang, 2015a; Golatkar et al., 2020; Izzo et al., 2021) propose sharding data into shards and retraining. However, these methods cannot be directly applied to graphs.
\subsection{Machine Unlearning in Graph Networks}
In recent years, significant advancements have been made in the field of machine unlearning across various domains such as image classification~\cite{tarun2021fast,chundawat2023can,chundawat2022zero}, regression~\cite{tarun2023deep}, federated learning~\cite{wu2022federated}, graph learning~\cite{chen2022graph,wang2023inductive,cong2023efficiently}, and more. Prior works on graph unlearning (Table~\ref{tab:relatedwork}) can be broadly categorized into the following categories. 

\textbf{Sharding and aggregation-based methods.} 
Applying \textsc{Sisa}~\cite{bourtoule2021machine} directly to graph data can destroy the graph's structural information and diminish model performance. To address this, \textsc{GraphEraser}~\cite{chen2022graph} uses two sharding algorithms that preserve the graph structure and node features, respectively. The first algorithm leverages community detection to partition the graph into shards while retaining its structural characteristics. The second algorithm transforms node features into embedding vectors, safeguarding their information. Additionally, a learnable aggregator combines the predictions of these shard models to enhance the significance of individual shards, contributing to the final model's performance.
%Although it defines edge unlearning, it does not explicitly support it. % Says GNNDelete
\textsc{Guide} \cite{wang2023inductive} extends \textsc{GraphEraser} from transductive to inductive graph settings. In the inductive context, where the graph can evolve, and test graph information remains concealed during training, the sharding algorithm incorporates sub-graph repairing to restore lost information. Furthermore, an enhanced similarity-based aggregator independently calculates the importance score of each shard.

These methods are not suitable for graph datasets for various reasons. Sharding disrupts the graph structure, leading to the loss of intricate relationships and dependencies. Poor data heterogeneity and insufficient training samples within each shard result in a notable degradation of GNN performance \cite{koch2023no}. Sharding adds additional computational costs to the training step, which escalate with the number of shards \cite{ramezani2021learn}, necessitating additional hyper-parameter tuning. The inference step requires aggregation methods based on the shard's importance score, affecting its performance and elevating costs further. These methods can only work if they have complete training data, prior knowledge about the training procedure, optimizations used, and more.

\textbf{Model-intrinsic methods.} 
To address the overhead costs associated with sharding and aggregation, these methods propose unlearning tailored to specific GNNs. \textsc{GraphEditor} \cite{cong2022grapheditor} achieves closed-form computation of the influence of deleted nodes using ridge regression, ensuring precise unlearning through theoretical guarantees. The exact unlearning method is versatile, accommodating scenarios like node and edge updates and introducing sub-graph sampling to manage scalability concerns. The authors extend it to \textsc{GraphProjector} \cite{cong2023efficiently} to unlearn node features by projecting the weight parameters of the pre-trained model onto a subspace that is irrelevant to the features of the nodes to be forgotten.

Since these methods are often designed with specific GNN architectures and scenarios in mind, they are incompatible with popular nonlinear GNNs and knowledge graphs. Sub-graph sampling adds additional cost to the unlearning process. They do not eliminate the potential for unintended leakage of sensitive information.

\textbf{Model-agnostic methods.} 
Another line of thought is to approximate the unlearning process so as
to avoid sharding as well as dependence on specific types of models. \textsc{CertUnlearn} \cite{chien2022certified},~\cite{chien2023efficient} computes gradients of the loss function for parameters, adjusting them to reduce data point influence. Matrix computations refine updates based on parameter-data relationships, minimizing their impact on predictions. It provides upper bounds on certain norms of the gradient residuals, which gives a measure of confidence in the unlearning mechanism. To approximate unlearning, another recent work \textsc{Gif} \cite{wu2023gif} employs influence functions. It estimates the parameter changes in a GNN when specific graph elements are removed from the graph while considering the structural dependencies. This makes it better than traditional influence functions. \textsc{CertUnlearnEdge} \cite{wu2023certified} proposes certified unlearning specifically for edges introducing a linear noise term in the training loss to hide the gradient residual and using influence functions to estimate parameter changes.
Recently, \textsc{GNNDelete} \cite{cheng2023gnndelete} proposes a layer-wise deletion operator that removes the influence of deleted nodes and edges both from model weights and neighboring representations to preserve the remaining knowledge. 

Approximate methods face the challenge of effectively removing the influence of graph elements that share intricate relationships and dependencies. Information may still be implicitly encoded in retained elements making unlearning ineffective. However, extensive unlearning may undermine the performance of the retained elements due to a solid dependency on the deleted local graph neighborhoods. Further, the unlearned model must prevent unintended information leakage; success in membership inference attacks suggests lingering forgotten data traces even after unlearning.

Finally, a time-efficient unlearning method with minimal impact on training and minimal overhead costs in inference that closely approximates the performance of a retrained, or \textit{gold} model is needed.
% These being an approximate unlearning process can only provide an approximation of unlearning by adding random noise. Therefore, they lack a perfect guarantee for unlearning in practice \cite{thudi2022necessity}. 

\textbf{Other related graph unlearning works.}
While the works mentioned above focus on node and edge unlearning, \textsc{Sub-graphUnlearn}\cite{pan2023unlearning} focuses on unlearning sub-graph classification. It proposes a nonlinear graph learning framework based on Graph Scattering Transform. The goal is to make the model forget the representation of a specific target sub-graph, thereby obscuring its classification accuracy. 
\textsc{FedLU} \cite{zhu2023heterogeneous} presents a federated learning framework for unlearning knowledge graph embeddings in a heterogeneous setting. From cognitive neuroscience, it combines ideas of using retroactive interference and passive decay steps to recover performance and, thus, prevent forgotten knowledge from affecting the results. Federated learning involves training models on decentralized data sources, and therefore, it may raise concerns about data privacy and security
% \textcolor{red}{Seems like a generic unlearning technique that improves SISA, remotely connected to graph unlearning} 
Another recent work \textsc{Safe} \cite{dukler2023safe} highlights a trade-off between the cost of unlearning and the inference cost of the model. As the number of shards increases, unlearning cost decreases, but the inference cost increases. To address this, they propose the concept of a shard graph, which incorporates information from other shards at the time of training. Thus, the inference cost decreases significantly with a slight increase in the unlearning cost. Increasing information from other shards may increase training time per shard. If the number of shards is large, it could lead to longer training times and higher resource requirements. A detailed comparison of our proposed work with the existing state-of-the-art graph unlearning methods is presented in Table~\ref{tab:relatedwork}.
\begin{table*}[htb]
\centering
\resizebox{\textwidth}{!}{%
\begin{tabular}{@{}p{2.5cm}p{1.85cm}p{1.225cm}p{1.2cm}p{1.55cm}p{1.19cm}p{1.15cm}p{1.8cm}p{0.95cm}p{3cm}@{}}
\toprule
\textbf{Method} & \textbf{Model Paradigm} & \textbf{Impact on dataset} & \textbf{Training Overhead} & \textbf{Impact on performance} & \textbf{Inference Overhead} & \textbf{Membership Inference} & \textbf{Unlearning performance} & \textbf{Unlearning Time} & \textbf{Scalability} \\
\midrule
\textsc{GraphEraser}\cite{chen2022graph}, \textsc{Guide}\cite{wang2023inductive}                                             & Sharding \& aggregation based    & Disrupts graph structure & Sharding     & Poor performance on shards   & Aggregation         & Prone to data leaks           & Worst case: requires full retraining                                      & High   & Low, as the number of shards increases \\
\midrule
\textsc{GraphEditor}\cite{cong2022grapheditor}, \textsc{GraphProjector}\cite{cong2023efficiently}                                    & Model-intrinsic, linear GNNs only & None                     & None         & None                         & Sub-graph sampling  & Prone to MIA  & Limited                                                                           & Medium & Scales poorly with rising computation cost for forget set's influence \\
\midrule
\textsc{GNNDelete}\cite{cheng2023gnndelete}, \textsc{CertUnlearn-Edge}\cite{wu2023certified}, \textsc{GIF}\cite{wu2023gif}, \textsc{CertUnlearn}\cite{chien2022certified}\cite{chien2023efficient} & Model agnostic, applicable to all & None                     & Linear noise & Affected due to linear noise & Layer-wise deletion & Not as robust       & Ineffective with intricate node-edge links   & Medium & Matrix computation for parameter changes scales poorly \\
\midrule
\textbf{\textsc{D2DGN} \textit{(Ours)}}                                                                   & \textbf{Model agnostic, applicable to all} & \textbf{None}                     & \textbf{None}         & \textbf{None}                         & \textbf{None}                & \textbf{Robust to MIA} & \textbf{Effective in removal and preservation} & \textbf{Lowest} & \textbf{Scalable, effective even when graph size increases} \\
\bottomrule
\end{tabular}%
}
\caption{Comparison of Different Graph Unlearning Methods}
\label{tab:relatedwork}
\end{table*}

\subsection{Knowledge Distillation}
Knowledge distillation and its applications have aroused considerable attention in recent few years \cite{gou2021knowledge}. Various perspectives of knowledge, distillation schemes, teacher-student architectures \cite{wang2021knowledge}, distillation algorithms and applications have been studied \cite{papernot2016distillation}\cite{wang2019private}. Recently, knowledge distillation has been proposed as a promising technique for machine unlearning.~\cite{chundawat2023can} proposes a student-teacher architecture-based knowledge distillation technique where knowledge from competent and incompetent teachers is selectively transferred to the student. However, it is not directly applicable to graph datasets containing relationships and dependencies between elements. 
% Moved to supplementary

\begin{figure*}[htb]
\centering
\includegraphics[width=0.89\linewidth]{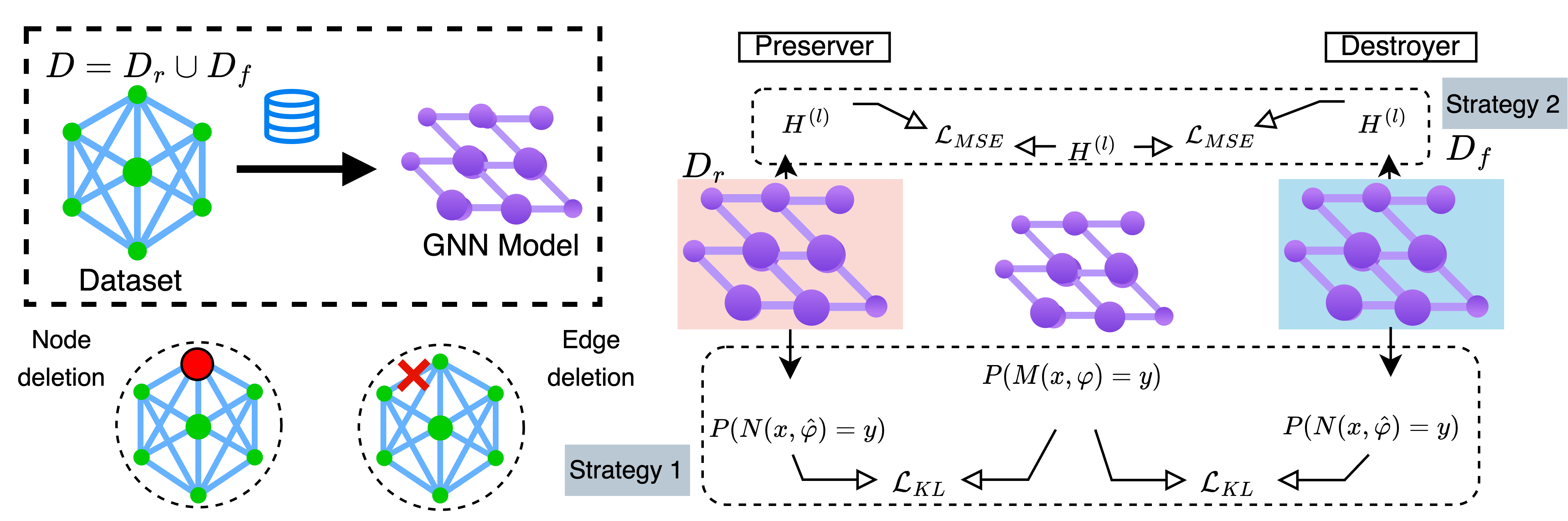}
\caption{This figure illustrates the proposed method. The \textit{GNN Model} is the original model trained on the complete data. The edge or node deletion requests are carried out as per the proposed \name. We have shown Strategy 1 and Strategy 2 of our work here.}
\label{fig:arch}
\end{figure*}

\section{Preliminaries}
\label{sec:method}
% \begin{figure}[t]
% \includegraphics[width=\linewidth]{fig/Method.pdf}
% \caption{Unlearning Overview: An unlearning algorithm takes a pre-trained model and selected instances from the training set as input to generate an updated model resembling the \textsc{Gold} model, trained without the forgotten set.}
% \label{fig:method}
% \end{figure}
\textbf{Graph Unlearning.} 
Let $\D=\{x_i, y_i\}_{i=1}^N$ represent a dataset of $N$ samples, where for every $i^{th}$ sample $x_i\in\mathbb{R}$, there is a corresponding output $y_i\in\mathbb{R}$. The aim is to forget a set of data points, represented by $\D_f$, while retaining another set of data points, represented by $\D_r$. It holds that $\D_r \bigcup \D_f = \D$ and $\D_r \bigcap \D_f = \emptyset$. Let $M(\cdot, \varphi)$ be a model with parameters $\varphi$. Given an input $x$, the model's output is $M(x,\varphi)$. For a machine learning algorithm $A$, it generates model parameters as $\varphi = A(\D)$. A \textit{gold model} is trained from scratch only on the retain set $\D_r$, denoted by $\varphi_r = A(\D_r)$. An unlearning algorithm $U$ utilizes all or a subset of $\D_r$ and $\D_f$, as well as the original model $\varphi$ to generate an unlearned model $\varphi_u$. Hence, 
\begin{align}
\label{eq:unlearningEqn}
    \varphi_u = U(\varphi, \D_r, \D_f).    
\end{align}
Extending this in the context of graphs, let $\G = (\V, \E)$ represent an attributed graph, where $\E$ denotes the set of $|\E|$ edges, and $\V$ denotes the set of $|\V|$ nodes. Each vertex $v \in \V$ is associated with a label denoted as $y_v$, and each edge $e \in \E$ has an associated label $y_e$. For edge unlearning, $\D =\{e_i, y_{e_i}\}_{i=1}^{|\E|}, e_i \in \E$. The set of edges marked for forgetting is $\D_f = \E_f \subseteq \E$, and the set of edges marked for retaining is $\D_r = \E_r = \E \backslash \E_f$. Similarly, for node unlearning, $\D = \{v_i, y_{v_i}\}_{i=1}^{|\V|}, v_i \in \V$, $\D_f = \V_f \subseteq \V$ and $\D_r = \V_r = \V \backslash \V_f$. 

% \textbf{Graph Neural Networks.} A GNN $M(\cdot, \varphi)$ is composed of several layers $g^{(1)}, \ldots, g^{(l)}$ which operates on graph $\G$. Each GNN layer $g^{(l)}$ is responsible for updating the node embeddings $\mathbf{H}^{(l)} = {\mathbf{h}_u^{(l)}, \ldots, \mathbf{h}_v^{(l)}}$, where $\mathbf{h}_u^{(l)}, \mathbf{h}_v^{(l)} \in \mathbb{R}^n$ represent the embeddings of nodes $u$ and $v$ at layer $l$, respectively. The GNN layer operates on the node embeddings and exchanges information with neighboring nodes to capture complex relationships and dependencies within the graph data.

% After training a GNN on the graph data is complete, the three-fold inference step involves obtaining final node embeddings $\mathbf{H}^{(l)}$, calculating logits (unnormalized probabilities) $\mathbf{Z}^{(l)}$ for tasks and applying the softmax function to the logits to obtain the probabilities $\mathbf{P}^{(l)}$:
% \begin{equation}
% \label{eq:gnninf}
% p_{ui}^{(l)} = \frac{\exp(z_{ui}^{(l)}/T)}{\sum_{j=1}^C \exp(z_{uj}^{(l)}/T)}
% \end{equation}
% where $p_{ui}^{(l)}$ is the probability that node $u$ belongs to class $i$, $T$ is the temperature parameter to control the sharpness of the probability distribution, and $C$ is the number of classes.
\textbf{Problem Formulation.}
Given an attributed graph dataset $\G$ and a GNN model $M(\cdot, \varphi)$,  devise an unlearning algorithm $U$, that unlearns the forget set $\D_f$ to obtain an unlearned model $M(\cdot, \varphi_u)$ with updated parameters $\varphi_u = U(\varphi, \D_r, \D_f)$, such that $\varphi_u$ closely approximates the performance of the gold model:
\begin{align}
    P(M(x, \varphi_u)=y) \approx P(M(x, \varphi_r)=y),\quad \forall x \in \D
\end{align}
where $P(X)$ denotes the probability distribution of any random variable $X$. 
Close approximation requires that the unlearned model possesses the following set of crucial properties:

\textit{Consistency}: The unlearning algorithm $U$ should effectively remove the influence of entities in $\D_f$. The parameters $\varphi_u$ should lead to a probability distribution that closely aligns with the gold model's probability distribution. $P(M(x, \varphi_u)=y) - P(M(x, \varphi_r)=y) \leq \epsilon_f, \forall x \in \D_f.$
% \begin{align}
%     P(M(x, \varphi_u)=y) - P(M(x, \varphi_r)=y) \leq \epsilon_f, \forall x \in \D_f.
% \end{align}
If $\epsilon_f$ is high, the unlearning process is not very successful in reducing the influence of the forget set. The impact of the forget set still holds significant sway over the model's predictions.
Conversely, if $\epsilon_f$ is low, it suggests a potential overfitting issue on the forget set due to extensive unlearning. The influence still exists, which the model uses to misclassify the forget set deliberately. So, $\epsilon_f$ should tend to zero.

\textit{Integrity}: Equally important is preserving knowledge from $\D_r$. $P(M(x, \varphi_u)=y) - P(M(x, \varphi_r)=y) \leq \epsilon_r, \forall x \in \D_r.$
% \begin{align}
%     P(M(x, \varphi_u)=y) - P(M(x, \varphi_r)=y) \leq \epsilon_r, \forall x \in \D_r.
% \end{align}
A high value of $\epsilon_r$ indicates that the model's generalization is compromised, as it becomes excessively tailored to the specific traits of the retain set.
On the other hand, if $\epsilon_r$ is low, it indicates that the unlearning process has inadvertently led to the loss of characteristics of the retain set. So, $\epsilon_r$ should tend to zero.

\textit{Membership Privacy:} 
The unlearned model must refrain from inadvertently disclosing or leaking information regarding specific data points in its training dataset. If the model's predictions concerning membership achieve high accuracy, this could signal potential privacy vulnerabilities, suggesting that traces of the training data persist within the model even post-unlearning.

\textit{Efficiency:} The unlearning process should be faster than retraining the model from scratch. The unlearning method should be lightweight, generic and agnostic to model-specific architecture details. These qualities minimize prerequisites for training knowledge and optimization techniques while not necessarily requiring access to the complete training data.    

% \textit{Generality:} The unlearning algorithm should accommodate various unlearning requests and cater to diverse downstream tasks.

%The \textit{preserver} retains knowledge by distilling response-based soft targets and feature-based node embeddings from the retained set $\D_r$, utilizing KL Divergence and Mean Squared Error losses. The \textit{destroyer} facilitates targeted knowledge removal via distillation on the forget set $\D_f$. Together, they transform the original model into the unlearned model.

\section{Proposed Method}
%In the context of distillation, a liquid mixture, known as the source, is heated to produce vapor, representing knowledge, and then condensed back to obtain desired liquids, referred to as the sink.~\cite{hinton2015distilling} extend this fundamental concept to knowledge distillation, wherein refined knowledge from a larger, more complex model, the \textit{source}, is transferred to a smaller model, the \textit{sink}.
The proposed method \name departs from vanilla knowledge distillation~\cite{hinton2015distilling} to facilitate effective graph unlearning. Since separating the graph entities present in the retain and the forget sets is challenging due to their intricate relationships and dependencies, we introduce two models, or separators, that separate retained and deleted knowledge. The knowledge \textit{preserver} aids in retaining the knowledge, preventing its loss during unlearning. Conversely, the knowledge \textit{destroyer} aids in forgetting the targeted knowledge. \name encompasses three key aspects: distillation architecture, distillation measure, and knowledge type.

\textbf{Distillation architecture} is shown in Figure \ref{fig:arch}, which is composed of three key components:
The \textit{source}, $M(\cdot, \varphi)$, represents the larger, complex model for knowledge transfer. It is trained fully on the complete dataset by algorithm $A$, where $\varphi = A(\D)$. It operates in training mode. Knowledge \textit{preserver}, $M(\cdot, \varphi^*)$, aids to retain knowledge by supplying positive knowledge, or embeddings of connected nodes. 
% Positive knowledge refers to embeddings of connected nodes in the trained model. 
It has the same model parameters as the source, $\varphi^* = A(\D)$, but acts in inference mode. 
Knowledge \textit{destroyer} aids in forgetting targeted knowledge by either supplying neutral knowledge or negative knowledge. It acts in inference mode. 
Neutral knowledge refers to embeddings for connected nodes in the graph, but from a randomly initialized, untrained model, $N(\cdot, \hat{\varphi})$.
Negative knowledge refers to embeddings of unconnected nodes in the trained model, $N(\cdot, \overline{\varphi})$.
Finally, the \textit{unlearned model} is the transformed state of the source after distillation with the preserver and destroyer. It is denoted with updated parameters as $M(\cdot, \varphi_u)$.

We propose unlearning algorithm \name such that $\varphi_u = \name(\varphi, \varphi^*, \hat{\varphi}, \D_r, \D_f)$. Notice that as compared to Eq.\ref{eq:unlearningEqn}, it has two additional inputs, $\varphi^*$ and $\hat{\varphi}$.
% \name trains source $M(\cdot, \varphi)$ further to transform it into the sink $M(\cdot, \varphi_u)$, with  such that it
To preserve knowledge, \name minimizes the loss between the source and preserver for the retain set $\D_r$. 
\begin{align}
\label{eq:loss_r}
\begin{split}
    \text{Loss}_r &= \mathcal{L}\;\Big(M(x,\varphi)\;, \;M(x,\varphi^*)\Big), \forall x \in \D_r
\end{split}
\end{align}
Similarly, to destroy targeted knowledge, \name minimizes loss between the source and destroyer for the forget set $\D_f$. Two options exist for the knowledge destroyer, to either supply neutral or negative knowledge:
\begin{align}
\label{eq:loss_f1}
\begin{split}
    \text{Loss}_f &= \mathcal{L}\;\Big(M(x,\varphi)\;, \;N(x,\hat{\varphi})\Big), \forall x \in \D_f  
\end{split}
\end{align}
\begin{align}
\label{eq:loss_f2}
\begin{split}
    \text{Loss}_f &= \mathcal{L}\;\Big(M(x,\varphi)\;,\;N(x,\overline{\varphi})\Big), \forall x \in \D_f 
\end{split}
\end{align}
where $\mathcal{L}$ represents the chosen distillation measure.\par

\textbf{Distillation measure} refers to the specific loss used to refine the knowledge of the source to unlearn. We explore two loss measures to achieve this: Kullback-Leibler Divergence (KL-divergence) \cite{kullback1951information} and Mean Squared Error (MSE). 

The KL-divergence measures the similarity between two probability distributions. For two models $M(\cdot, \varphi)$ and $N(\cdot, \varphi')$, the KL-divergence is defined by
{
\begin{align}
    \mathcal{L}_{\text{KL}}\;\Big(M(\cdot,\varphi)\;||\;N(\cdot,\varphi')\Big)
    &= E_{x \sim P(x)}\left[\log \frac{P(M(x, \varphi)=y)}{P(N(x, \varphi')=y)}\right]
\end{align}}
$\forall x \in \D$.
The Mean Squared Error (MSE) loss function between the two sets of node embeddings can be formally defined as follows:
% \begin{align}
%     \ell_{\text{MSE}}\Big(\mathbf{h}_u^{(l)}, \mathbf{h}_v^{(l)}\Big) = \frac{1}{d} \sum_{i=1}^{d} (h_{u,i}^{(l)} - h_{v,i}^{(l)})^2
% \end{align}
% where, $\mathbf{h}_u^{(l)}$ and $\mathbf{h}_v^{(l)}$ represent the node embeddings of nodes $u$ and $v$ at GNN layer $l$, respectively.
% $d$ is the number of dimensions of the embeddings.
% $h_{u,i}^{(l)}$ and $h_{v,i}^{(l)}$ denote the $i$-th component of the embeddings for nodes $u$ and $v$ at layer $l$, respectively. The loss function for the entire matrix of node embeddings at layer $l$ is given by:
% \begin{align}
%     \mathbf{l}_{\text{MSE}}\Big(\mathbf{H}_{M}^{(l)}, \mathbf{H}_{N}^{(l)}\Big) = \frac{1}{n^2} \sum_{u,v} \ell_{\text{MSE}}\Big(\mathbf{h}_u^{(l)}, \mathbf{h}_v^{(l)}\Big)
% \end{align}
% where, $\mathbf{H}_{M}^{(l)}$ and $\mathbf{H}_{N}^{(l)}$ are the matrices of node embeddings for models $M(\cdot, \varphi)$ and $N(\cdot, \varphi')$ at layer $l$, respectively. Finally, the overall loss function that sums the MSE over all layers is given by:
\begin{align}
    \mathcal{L}_{\text{MSE}}\Big(M(\cdot, \varphi), N(\cdot, \varphi')\Big) = \sum_{l} \textbf{l}_{\text{MSE}}\Big(\mathbf{H}_{M}^{(l)}, \mathbf{H}_{N}^{(l)}\Big)
\end{align}
where $\mathbf{H}_{M}^{(l)}$ and $\mathbf{H}_{N}^{(l)}$ are the matrices of node embedding for models $M(\cdot, \varphi)$ and $N(\cdot, \varphi')$ at layer $l$, respectively. These measures evaluate the similarity between different knowledge types. %soft targets and feature embeddings, respectively. 

\textbf{Knowledge type} refers to the different forms in which the source $M(\cdot, \varphi)$ stores knowledge, which can be distilled.
\textit{Response-based knowledge} involves capturing the neural response of the last output layer of the source. The main objective is to mimic the final predictions of the source directly. The knowledge source generates soft targets, representing probabilities that the input belongs to different classes. These probabilities can be estimated using the softmax function, as shown in Eq. \ref{eq:gnninf}: 
\begin{equation}
\label{eq:gnninf}
p_{ui}^{(l)} = \frac{\exp(z_{ui}^{(l)}/T)}{\sum_{j=1}^C \exp(z_{uj}^{(l)}/T)}
\end{equation}
where $p_{ui}^{(l)}$ is the probability that node $u$ belongs to class $i$, $T$ is the temperature parameter to control the sharpness of the probability distribution, and $C$ is the number of classes.
\textit{Feature-based knowledge} involves capturing multiple levels of feature representations from the intermediate layers of the source. The primary objective is to directly match the feature activation of the source and the separator. \name utilizes node embeddings to effectively capture essential structural information and patterns from the source. It alters the GNN architecture to extract embedding from all layers, rather than just the last layer, during the forward pass, i.e., $\mathbf{H}^{(1)}, \ldots, \mathbf{H}^{(l)}$, where $l$ is the total number of layers in the GNN. 

\textbf{Strategies}. We present three unlearning strategies combining the distillation architecture, knowledge type, and distillation measures. Overall, the unlearning objective is formulated in Eq.\ref{eq:loss} as follows:
\begin{align}
\label{eq:loss}
\begin{split}
    \text{Loss} &= \alpha \cdot \text{Loss}_r + (1-\alpha) \cdot \text{Loss}_f
\end{split}
\end{align}
where $\alpha$ is the regularization coefficient, that balances the trade-off between the effects of the two separators.

\textit{Strategy 1} employs response-based knowledge, utilizing soft targets for minimizing the KL divergence loss. The destroyer is $N(\cdot, \hat{\varphi})$.
\begin{align}
\begin{split}
    \text{Loss} &= \alpha \cdot \mathcal{L}_{\text{KL}}\;\Big(M(x,\varphi)\;||\;M(x,\varphi^*)\Big)\\ &+ (1-\alpha)\cdot \mathcal{L}_{\text{KL}}\;\Big(M(x',\varphi)\;||\;N(x',\hat{\varphi})\Big)\\ &, \forall x \in \D_r, \forall x' \in \D_f
\end{split}
\end{align}
\textit{Strategy 2} employs feature-based knowledge, utilizing the neutral knowledge, minimized with MSE. The destroyer is $N(\cdot, \hat{\varphi})$. \textit{Strategy 3} is the same as \textit{Strategy 2} but uses negative knowledge $N(\cdot, \overline{\varphi})$ as the destroyer.
\begin{align}
\begin{split}
    \text{Loss} &= \alpha \cdot \mathcal{L}_{\text{MSE}}\Big(M(x, \varphi), M(x, \varphi^*)\Big)\\ 
    &+ (1-\alpha) \cdot \mathcal{L}_{\text{MSE}}\Big(M(x', \varphi), N(x', \hat{\varphi})\Big)\\ &, \forall x \in \D_r, \forall x' \in \D_f
\end{split}
\end{align}
\begin{align}
\begin{split}
    \text{Loss} &= \alpha \cdot \mathcal{L}_{\text{MSE}}\Big(M(x, \varphi), M(x, \varphi^*)\Big)\\ 
    &+ (1-\alpha) \cdot \mathcal{L}_{\text{MSE}}\Big(M(x', \varphi), N(x', \overline{\varphi})\Big)\\ &, \forall x \in \D_r, \forall x' \in \D_f
\end{split}
\end{align}

\textbf{Information bound.} 
% blindspot = knowledge destroyer
% retrained = knowledge preserver
% unlearned = source
\begin{figure}[tb]
\centering
\includegraphics[width=\columnwidth]{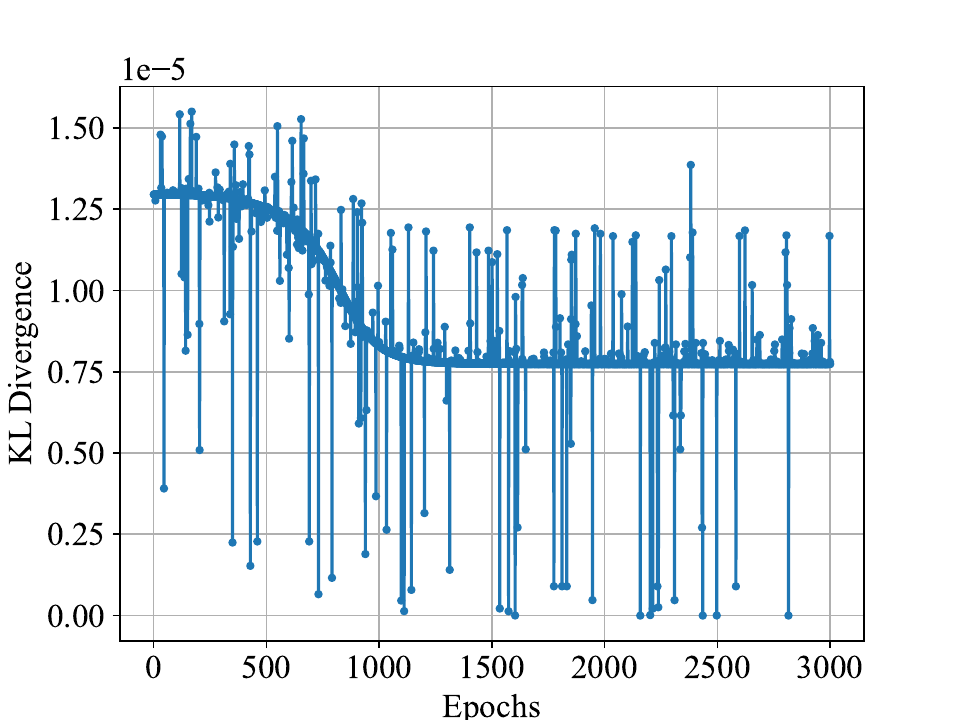}

\caption{KL-Divergence between knowledge destroyer and knowledge preserver with respect to the increasing number of epochs. We observe that with increasing epochs, the knowledge destroyer is reaching closer to the prediction distribution of the knowledge preserver on the forget set $\D_f$.}
\label{fig:infoBound}
\end{figure}
Information about the forget set in the unlearned model is bounded by the information contained in the knowledge destroyer. In their work, \cite{golatkar2020eternal} employ read-out functions and leverage KL-Divergence between the distributions obtained from the unlearned and retrained models for the forget set. This serves as a metric for the remaining information in classification problems. In our graph context, we use the identity function as our read-out function, we use the predicted values themselves for comparing distributions. We denote the information within a model $M(\cdot,\varphi)$ regarding a dataset $\D$ as $\mathcal{I}(M(\D,\varphi))$. Then, 
\begin{align}
\mathcal{I}(M(\D_f,\varphi)) &\approx \mathcal{I}(N(\D_f,\hat{\varphi}))
\end{align}
\begin{align}
or, \mathcal{I}(N(,\hat{\varphi}), \D_f) &\propto \mathcal{L}_{\text{KL}}\;\Big(M(x,\varphi)\;, \;N(x,\hat{\varphi_r})\Big)
\end{align}
\begin{align}
or, \mathcal{I}(N(,\hat{\varphi}), \D_f) &= k \cdot \mathcal{L}_{\text{KL}}\;\Big(M(x,\varphi)\;, \;N(x,\hat{\varphi_r})\Big)
\end{align}
where $k$ is a constant of proportionality. 

In Figure~\ref{fig:infoBound}, we depict a graph illustrating the KL-Divergence (between the knowledge destroyer and knowledge preserver) as number of epochs increase. Notice that the knowledge destroyer reaches closer to the prediction distribution of the knowledge preserver on the forget set. 
This convergence can be expressed as $\mathcal{L}_{\text{KL}}\;\Big(M(x,\varphi);, ;N(x,\hat{\varphi_r})\Big) \leq \epsilon$, where $\epsilon \propto 1/n$ and $n$ represents the epochs for which the knowledge destroyer is trained. If $\epsilon = c/n$, with $c$ being a constant, then $\mathcal{L}_{\text{KL}}\;\Big(M(x,\varphi);, ;N(x,\hat{\varphi_r})\Big) \leq kc/n$. Consequently, the disclosed information by \name is constrained by $kc/n$.
% In Figure~\ref{fig:infoBound}, we plot a graph to show the KL-Divergence (between knowledge destroyer and  knowledge preserver) with respect to the increasing number of epochs. This can be denoted as $ \mathcal{L}_{\text{KL}}\;\Big(M(x,\varphi)\;, \;N(x,\hat{\varphi_r})\Big) \leq \epsilon$ and $\epsilon \propto 1/n$ where $n$ denotes the epochs for which knowledge destroyer is trained. If $\epsilon = c/n$, where $c$ is a constant, then $\mathcal{L}_{\text{KL}}\;\Big(M(x,\varphi)\;, \;N(x,\hat{\varphi_r})\Big) \leq kc/n$. Therefore, The amount of information \name reveals is bounded by $kc/n$.

\textbf{Theoretical analysis of time complexity.}
The GNN forward pass has the time complexity $O(L \cdot N \cdot F^2 + L \cdot |E| \cdot F)$, where $N$, $F$, $|E|$ and $L$ are the number of nodes, node features, edges and GNN layers, respectively. The complexity of computation of distillation losses are, KL divergence: $O(N\cdot d)$ and Mean Squared Error: $O(N\cdot d^2)$, where $d$ is dimensionality of node embeddings. The GNN forward pass is the dominant factor whereas computation of distillation losses is a smaller factor. Hence, the time complexity of the proposed method is $O(L \cdot N \cdot F^2 + L \cdot |E| \cdot F)$.

\begin{table}[]
\centering
\caption{Statistics of evaluated datasets}
\label{tab:datasets}
\begin{tabular}{@{}lll@{}}
\toprule
Dataset        & \#Nodes & \#Edges  \\ \midrule
CiteSeer       & $3327$    & $9104$     \\
Cora           & $19793$   & $126842$   \\
CS             & $18333$   & $163788$   \\
DBLP           & $17716$   & $105734$   \\
PubMed         & $19717$   & $88648$    \\
% Physics        & 34493   & 247962   \\
% ogbl-collab    & 253868  & 2358104  \\
% ogbl-citation2 & 2927963 & 15228622 \\ 
\bottomrule
\end{tabular}
\end{table}
\begin{table*}[t]
\centering
\caption{AUC results for link prediction when unlearning 2.5\% edges $\E_f = \E_{f,\mathrm{IN}}$ on DBLP dataset. The closest method to \textsc{Gold} model is marked in \textbf{bold}, and the second closest is \underline{underlined}. `-' denotes that method is not applicable for those GNNs.}
\label{tab:AcrossGNNsAcc}
\resizebox{\linewidth}{!}{%
\begin{tabular}{@{}lcccccc@{}}
\toprule
\multicolumn{1}{c}{}                        & \multicolumn{2}{c}{GCN}       & \multicolumn{2}{c}{GAT}       & \multicolumn{2}{c}{GIN} \\ \cmidrule(l){2-7}
\multicolumn{1}{c}{\multirow{-2}{*}{Model}} & \multicolumn{1}{c}{$\E_r$}    & \multicolumn{1}{c}{$\E_f$}    & \multicolumn{1}{c}{$\E_r$}                   & \multicolumn{1}{c}{$\E_f$}    & \multicolumn{1}{c}{$\E_r$}    & \multicolumn{1}{c}{$\E_f$} \\ \midrule
\rowcolor[HTML]{EFEFEF} 
\textsc{Gold}                               & $0.964 \pm 0.003$             & $0.506 \pm 0.013$             & $0.956 \pm 0.002$                            & $0.525 \pm 0.012$             & $0.931 \pm 0.005$             & $0.581 \pm 0.014$ \\
\textsc{GradAscent}                         & $0.555 \pm 0.066$             & $0.594 \pm 0.063$             & $0.501 \pm 0.020$                            & $0.700 \pm 0.025$             & $0.524 \pm 0.017$             & $\underline{0.502 \pm 0.002}$ \\
\textsc{D2D}                                & $0.500 \pm 0.000$             & $\underline{0.500 \pm 0.000}$ & $0.500 \pm 0.000$                            & $\underline{0.500 \pm 0.000}$ & $0.500 \pm 0.000$             & $0.500 \pm 0.000$ \\
\textsc{GraphEraser}                        & $0.527 \pm 0.002$             & $\underline{0.500 \pm 0.000}$ & $0.538 \pm 0.013$                            & $\underline{0.500 \pm 0.000}$ & $0.517 \pm 0.009$             & $0.500 \pm 0.000$ \\
\textsc{GraphEditor}                        & $0.776 \pm 0.025$             & $0.432 \pm 0.009$             & -                                            & -                             & -                           & - \\
\textsc{CertUnlearn}                        & $0.718 \pm 0.032$             & $0.475 \pm 0.011$             & -                                            & -                             & -                           & - \\
% \textsc{GIF}                          & $\underline{0.934 \pm 0.002}$ & $0.748 \pm 0.006$             & $\underline{0.914 \pm 0.007}$                & $0.774 \pm 0.015$             & $\underline{0.897 \pm 0.006}$ & $0.740 \pm 0.015$ \\
\textsc{GNNDelete}                          & $\underline{0.934 \pm 0.002}$ & $0.748 \pm 0.006$             & $\underline{0.914 \pm 0.007}$                & $0.774 \pm 0.015$             & $\underline{0.897 \pm 0.006}$ & $0.740 \pm 0.015$ \\
\name (\textit{Ours})                       & $\mathbf{0.958 \pm 0.009}$    & $\mathbf{0.501 \pm 0.103}$    & $\mathbf{0.939 \pm 0.019}$                   & $\mathbf{0.534 \pm 0.010}$    & $\mathbf{0.937 \pm 0.014}$    & $\mathbf{0.541 \pm 0.011}$ \\
\bottomrule
\end{tabular}
}
\end{table*}
\begin{table}[t]
\centering
\caption{MI Ratio(↑) results for link prediction when unlearning 2.5\% edges $\E_f = \E_{f,\mathrm{IN}}$ on DBLP dataset. The best method is marked in \textbf{bold}, and the second best is \underline{underlined}.}
\label{tab:AcrossGNNsMI}
% \resizebox{\linewidth}{!}{%
\begin{tabular}{@{}llll@{}}
\toprule
\multicolumn{1}{c}{Model} & \multicolumn{1}{c}{GCN} & \multicolumn{1}{c}{GAT} & \multicolumn{1}{c}{GIN} \\ \midrule
\rowcolor[HTML]{EFEFEF} 
\textsc{Gold} & $1.255 \pm 0.207$ & $1.223 \pm 0.151$ & $1.200 \pm 0.177$ \\
\textsc{GradAscent} & $1.180 \pm 0.061$ & $1.112 \pm 0.109$ & $1.123 \pm 0.103$ \\
\textsc{D2D} & $1.264 \pm 0.061$ & $1.112 \pm 0.109$ & $1.123 \pm 0.103$ \\
\textsc{GraphEraser} & $1.101 \pm 0.032$ & $1.264 \pm 0.000$ & $\underline{1 . 2 6 4 \pm 0 . 0 0 0}$ \\
\textsc{GraphEditor} & $1.189 \pm 0.193$ & $1.189 \pm 0.104$ & $1.071 \pm 0.113$ \\
\textsc{CertUnlearn} & $1.103 \pm 0.087$ & $1.103 \pm 0.087$ & $1.103 \pm 0.193$ \\
\textsc{GNNDelete} & $\underline{1.266 \pm 0.106}$ & $\underline{1.338 \pm 0.122}$ & $1.254 \pm 0.159$ \\
\name (\textit{Ours}) & $\mathbf{2.531 \pm 0.163}$          & $\mathbf{1.715 \pm 0.108}$                   & $\mathbf{8.485 \pm 0.181}$                   \\ \bottomrule
\end{tabular}
% }
\end{table}
\begin{table*}[t]
\centering
\caption{AUC performance on link prediction when unlearning 
$2.5\%$ edges $\E_f = \E_{f,\mathrm{IN}}$ and $\E_{f,\mathrm{OUT}}$ across multiple datasets: CiteSeer \protect\cite{yang2016revisiting}, Cora \protect\cite{yang2016revisiting}, CS \protect\cite{shchur2018pitfalls}, DBLP \protect\cite{fu2020magnn}, and PubMed \protect\cite{yang2016revisiting}. The closest method to \textsc{Gold} model is marked in \textbf{bold}.}
\resizebox{\linewidth}{!}{%
\begin{tabular}{@{}l>{\columncolor[HTML]{EFEFEF}}l >{\columncolor[HTML]{EFEFEF}}l llll|>{\columncolor[HTML]{EFEFEF}}l >{\columncolor[HTML]{EFEFEF}}l llll@{}}\toprule & \multicolumn{2}{l}{\cellcolor[HTML]{EFEFEF}\textsc{Gold}} & \multicolumn{2}{l}{\textsc{GNNDelete}} & \multicolumn{2}{l|}{\name} & \multicolumn{2}{l}{\cellcolor[HTML]{EFEFEF}\textsc{Gold}} & \multicolumn{2}{l}{\textsc{GNNDelete}} & \multicolumn{2}{l}{\name} \\ \cmidrule(l){2-13}
\multirow{-2}{*}{Dataset} & $\E_{r, \mathrm{IN}}$                                     & $\E_{f, \mathrm{IN}}$                  & $\E_{r, \mathrm{IN}}$      & $\E_{f, \mathrm{IN}}$                                     & $\E_{r, \mathrm{IN}}$                  & $\E_{f, \mathrm{IN}}$                           & $\E_{r, \mathrm{OUT}}$ & $\E_{f, \mathrm{OUT}}$ & $\E_{r, \mathrm{OUT}}$ & $\E_{f, \mathrm{OUT}}$ & $\E_{r, \mathrm{OUT}}$ & $\E_{f, \mathrm{OUT}}$      \\ \midrule
CiteSeer                  & $0.951$                                                   & $0.522$                                & $0.926$                    & $0.717$                                                   & $\mathbf{0.937}$                                & $\mathbf{0.387}$                                         & $0.938$                & $0.706$                & $0.951$                & $0.896$                & $\mathbf{0.936}$                & $\mathbf{0.578}$       \\
Cora                      & $0.966$                                                   & $0.520$                                & $0.925$                    & $0.716$                                                   & $\mathbf{0.964}$                                & $\mathbf{0.511}$                                         & $0.966$                & $0.790$                & $0.953$                & $0.912$                & $\mathbf{0.964}$                & $\mathbf{0.755}$       \\
CS                        & $0.968$                                                   & $0.545$                                & $0.941$                    & $0.731$                                                   & $\mathbf{0.965}$                                & $\mathbf{0.533}$                                         & $0.968$                & $0.767$                & $0.951$                & $0.897$               & $\mathbf{0.967}$                & $\mathbf{0.732}$       \\
DBLP                      & $0.964$                                                   & $0.506$                                & $0.934$                    & $0.748$                                                   & $\mathbf{0.958}$                                & $\mathbf{0.501}$                                         & $0.965$                & $0.777$                & $0.957$                & $0.892$                & $\mathbf{0.965}$                & $\mathbf{0.734}$       \\
PubMed                    & $0.968$                                                   & $0.499$                                & $0.920$                    & $0.739$                                                   & $\mathbf{0.969}$                                & $\mathbf{0.525}$                                         & $0.967$                & $0.696$                & $0.954$                & $0.909$                & $\mathbf{0.969}$                & $\mathbf{0.666}$       \\ \bottomrule
\end{tabular}%
}
\label{tab:AcrossDatasetsAcc}
\end{table*}
\section{Experimental Setup}
\label{sec:exp}
All the experiments are performed on 4x NVIDIA RTX2080 (32GB). We show the outcome of \textit{Strategy 1} in results, while a comprehensive comparison of the different strategies is provided in the ablation studies. The regularization coefficient $\alpha$ in Eq.~\ref{eq:loss} is set to $0.5$ in all experiments. We use the identical settings as in the case of \textsc{GNNDelete} and present their results exactly as reported in their paper.\par 

\textbf{Datasets.} We evaluate \name on several datasets: CiteSeer \cite{yang2016revisiting}, Cora \cite{yang2016revisiting}, CS \cite{shchur2018pitfalls}, DBLP \cite{fu2020magnn}, and PubMed \cite{yang2016revisiting}. The specific details of each dataset are presented in Table \ref{tab:datasets}.

\textbf{GNN Architectures used.} We comprehensively evaluate the flexibility and effectiveness of \name by testing it on three distinct GNN architectures. Specifically, we evaluate our approach on Graph Convolutional Networks (GCN)~\cite{kipf2016semi}, Graph Attention Networks (GAT)~\cite{velickovic2017graph}, and Graph Isomorphism Networks (GIN)~\cite{xu2018how}.
% For heterogeneous graphs, we assess our method on Relational Graph Convolutional Network (R-GCN) as presented by Schlichtkrull et al. (2018) and Relational Graph Attention Network (R-GAT) proposed by Chen et al. (2021b).

\textbf{Baselines and State-of-the-art.} The \textsc{Gold} model assumes a central role among the baseline methods. It serves as a benchmark by being trained exclusively on the retain set from scratch, against which the performance of the unlearned model is measured. Any other unlearned model is expected to closely approximate the \textsc{Gold} model's performance. Furthermore, our evaluation encompasses six additional baseline methods:
\textsc{GradAscent} employs gradient ascent on edge features using the cross-entropy loss, iteratively updating the model's parameters.
\textsc{D2D} Descent-to-delete \cite{neel2021descent} handles deletion requests sequentially using convex optimization and reservoir sampling.
\textsc{GraphEraser} uses partitioning and aggregation-based methods with optimizers to extend \textsc{Sisa} \cite{bourtoule2021machine} for graphs. 
\textsc{GraphEditor} \cite{cong2022grapheditor} is an approach that unlearns linear GNN models through fine-tuning based on a closed-form solution.
\textsc{CertUnlearn} \cite{chien2022certified} introduces a certified unlearning technique for linear GNNs.
\textsc{GNNDelete} \cite{cheng2023gnndelete} introduces a layer-wise deletion operator that removes the influence of deleted nodes and edges from model weights and neighboring representations while preserving the remaining knowledge.

\textbf{Evaluation Metrics.} 
We employ a diverse set of evaluation metrics to comprehensively assess the performance of \name. (a) \textit{AUC on the Forget set $\D_f$} measures how consistently the unlearned models can differentiate deleted elements (in $\D_f$) from the remaining elements (in $\D_r$). To calculate the area under the curve, we factor in the total count of the deleted elements within $\D_f$, and in the equal count, we sample retained elements from $\D_r$. Subsequently, we assign a label of 0 to the deleted elements and 1 to the retained elements. Values closer to the \textsc{Gold} model signify better unlearning \textit{consistency}, i.e., effectiveness in reducing the influence of deleted elements. (b) \textit{AUC on the Retain set $\D_r$} measures the prediction performance of the unlearned models. Values closer to the \textsc{Gold} model signify better unlearning \textit{integrity}, i.e., effectiveness in preserving the knowledge about retained elements. (c) \textit{Membership Inference (MI) Ratio}, derived from an MI attack~\cite{olatunji2021membership}, gauges the efficacy of unlearning by quantifying the probability ratio of $\D_f$ presence before and after the unlearning process. A ratio surpassing $1$ signifies reduced information retention about $\D_f$, indicating enhanced \textit{membership privacy}. Conversely, a ratio below $1$ suggests greater retention of information about $\D_f$. (d) \textit{Unlearning Cost} measures the cost of unlearning in the GNN. We present \textit{unlearning time}, representing the duration required for the unlearning process in a model. (e) \textit{Inference Cost} evaluates the overhead of the unlearning method on both training and inference stages of the original model. We quantify this impact by reporting the number of floating-point operations (FLOPs) executed by the unlearned model per forward pass.

\textbf{Sampling strategies. } We use two strategies for sampling edges in the forget set $\E_f$. When $\E_f = \E_{f,\mathrm{IN}}$, edges are sampled within the 2-hop enclosing sub-graph of $\E_r$. Conversely, when $\E_f = \E_{f,\mathrm{OUT}}$, edges are sampled outside the 2-hop enclosing sub-graph of $\E_r$.
% We propose two sampling strategies for edge unlearning tasks.
\section{Results}
\label{sec:results}

\subsection{Comparison with SOTA on GNN Architectures}
\label{sec:compare}
\textbf{Integrity:} We conduct a comparative analysis of \name against a range of state-of-the-art methods. Table \ref{tab:AcrossGNNsAcc} presents the results for AUC performance. On the retain set $\D_r$, for GCN architecture, \name consistently achieves performance levels closest to the \textsc{Gold} model, outperforming \textsc{GraphEraser}, \textsc{GraphEditor}, \textsc{CertUnlearn}, and \textsc{GNNDelete} by $43.1\%$, $18.2\%$, $24.0\%$, and $2.4\%$, respectively. Similarly, it outperforms other baselines across GNN architectures like GAT and GIN. These results substantiate that \name achieves the \textit{best integrity}, preserving critical knowledge through the unlearning process.\par

\textbf{Consistency:} On the forget set $\D_f$,  for GCN architecture, \name achieves performance levels closest to the \textsc{Gold} model, as close as $\delta=0.1\%$, outperforming \textsc{GraphEraser}, \textsc{GraphEditor}, \textsc{CertUnlearn}, and \textsc{GNNDelete} which achieve $\delta = 0.6\%$, $7.4\%$, $3.1\%$, and $24.2\%$, respectively. Similar results follow for GAT and GIN. These results substantiate that \name achieves the \textit{best consistency}, effectively removing knowledge through unlearning.\par
\textsc{GraphEraser} cannot capture the inherent intricate relationships and neighborhood dependencies since the graph structure is broken into shards and overfits specific shards. \textsc{GraphEditor} and \textsc{CertUnlearn}, constrained by their linear architectures, are not as effective. \textsc{Descent-to-delete} and \textsc{GradAscent} adjust the model parameters equally and do not adapt to the removal of specific data, thus losing their ability to unlearn.\par

It is intriguing to note that \textsc{GNNDelete} achieves an AUC significantly higher, seemingly better than even the \textsc{Gold} model. However, an overly elevated or diminished AUC may imply the potential presence of overfitting issues due to extensive unlearning. The influences from the forget set within the model persist, as reconfirmed by the results of the Membership Inference Attack.

\textbf{MI Ratio:} Table \ref{tab:AcrossGNNsMI} shows that \name outperforms all other baselines on the MI Ratio, highlighting its effectiveness in preventing leakage of forgotten data. For GCN architecture, it improves on the baselines \textsc{GraphEraser}, \textsc{GraphEditor}, \textsc{CertUnlearn}, and \textsc{GNNDelete} by $+1.43$, $+1.34$, $+1.42$, and $+1.26$, respectively. Similarly, it outperforms other baselines across other GNN architectures. These results substantiate that \name achieves the \textit{best membership privacy}, preventing leakage of forgotten data amidst the unlearning process. 

We evaluate \name on node unlearning as well. It involves the selective removal of individual nodes from a GNN without affecting the performance of the downstream tasks for the retained nodes. Table \ref{tab:NodeUnlearning} presents a comprehensive comparison of \name with the state-of-the-art models on the DBLP dataset. Noteworthy findings include the excellent performance of \textsc{GNNdelete}, marked in bold, with an AUC score closest to the \textsc{Gold} model with $\delta = 3.5\%$. Remarkably, \name achieves the second-best, competitive AUC score, denoted by underlining, with $\delta = 10.5\%$. 
% The slightly lower AUC score of \name compared to \textsc{GNNdelete} unveils challenges intrinsic to node unlearning. 
Due to similarities in properties and features among graph nodes, removing nodes can still implicitly retain features of deleted nodes. \textsc{GNNDelete}'s strategy of retaining node-specific representations in local graph neighborhood seems to be better than the global node feature-based distillation of \name. The trade-offs involved in preserving information during node and edge unlearning are delicate, affirming \name's competitive standing.

\subsection{Comparison with SOTA on Different Datasets}
Table~\ref{tab:AcrossDatasetsAcc} presents a performance of \name's performance across various datasets. While the datasets themselves differ significantly in terms of the number of nodes and edges, \name consistently showcases its effectiveness in unlearning across this wide spectrum of graph sizes and differing graph characteristics. Further, the results highlight the superiority of \name over \textsc{GNNDelete} for both the edge sampling strategies, $\E_{f,\mathrm{IN}}$ and $\E_{f,\mathrm{OUT}}$. Unlearning $\E_{f,\mathrm{IN}}$, which pertains to the edges within the 2-hop enclosing subgraph of $\E_r$, is challenging due to the intricate inter dependencies within the local neighborhood.\par

\subsection{Efficiency Analysis}

% \begin{figure}[t]
% \centering
% \includegraphics[width=0.35\textwidth]{fig/unlearning_time.png}
% \caption{Unlearning time comparison of our method with existing SOTA and Gold model for various datasets. \textit{Lower is better.}}
% \label{fig:unlearningtime}
% \end{figure}

\begin{figure}[tb]
\centering
% \begin{subfigure}{0.49\columnwidth}
% \centering
            \includegraphics[width=.85\columnwidth]{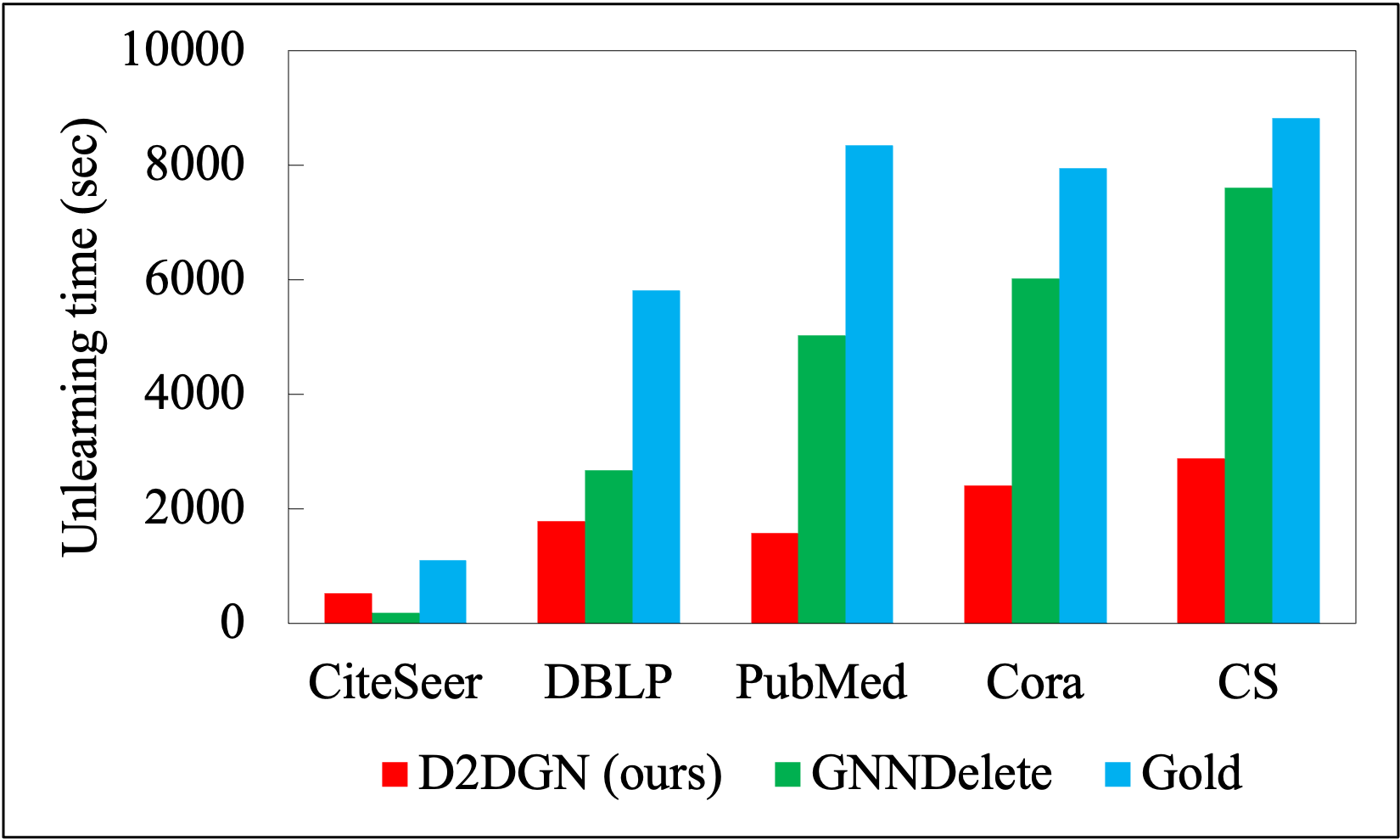}
            \caption{Unlearning time comparison across datasets: \name vs. SOTA and \textsc{Gold} models (↓).}
            \label{fig:unlearningtime}
% \end{subfigure}
% \begin{subfigure}{0.49\columnwidth}
% \centering
            \includegraphics[width=.85\columnwidth]{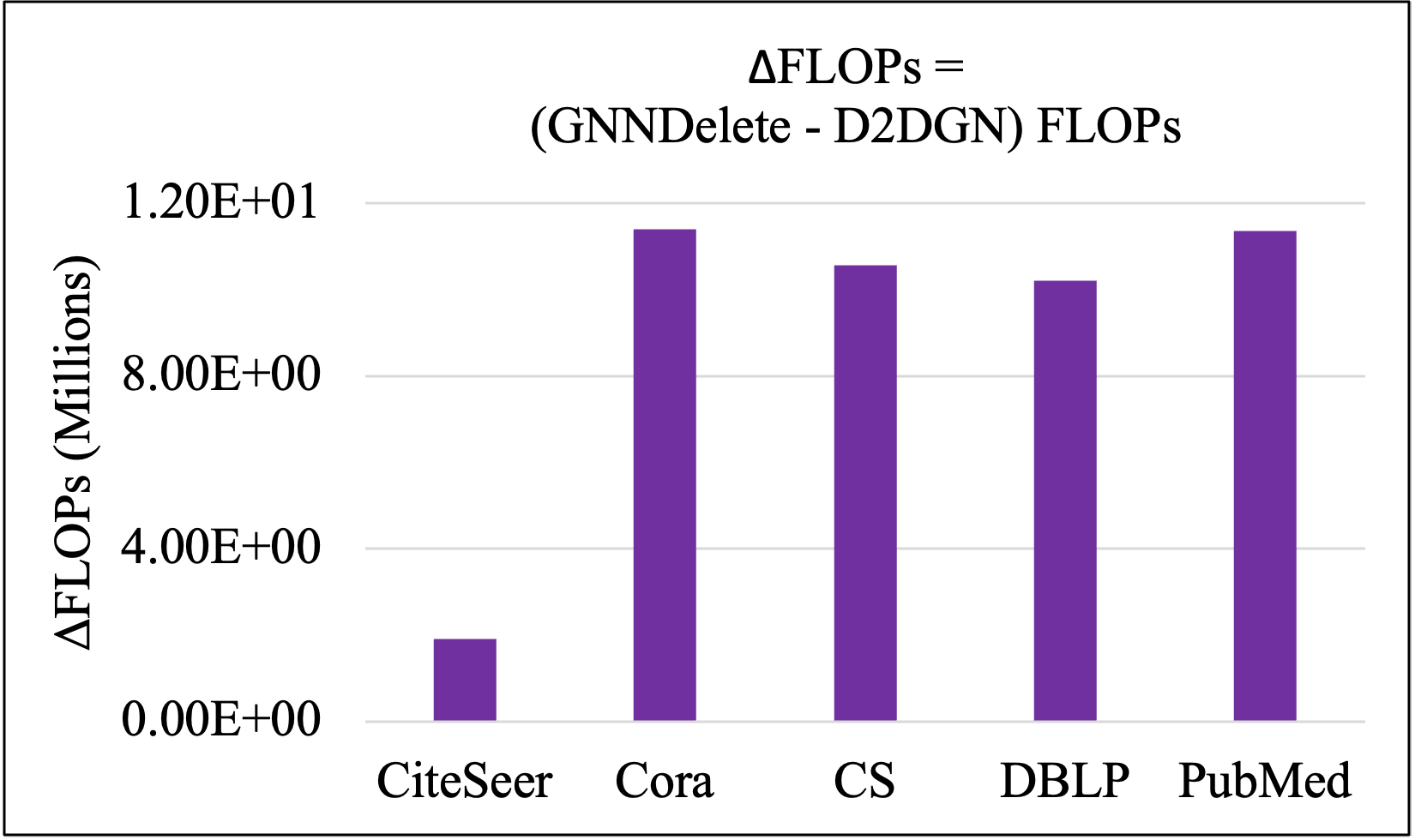}
            \caption{Difference in FLOPS: \name vs. \textsc{GNNDelete} across datasets (↓).}
            \label{fig:FLOPs}
% \end{subfigure}
\caption{Efficiency Analysis}
\label{fig:combined}
\end{figure}

\textbf{Unlearning time} comparison of our method with existing methods is shown in Figure~\ref{fig:unlearningtime}. \name demonstrates a remarkable speed advantage, being up to $\mathbf{5.3}\times$ faster than the \textsc{Gold} model. This is observed for all datasets with varying graph sizes and characteristics. Specifically, it is faster on datasets CiteSeer, DBLP, PubMed, Cora and CS by $2.1\times$, $3.2\times$, $5.3\times$, $3.2\times$ and $3.0\times$ respectively.
It is even faster than the existing state-of-the-art method \textsc{GNNDelete} for all datasets except for very small datasets like CiteSeer. Specifically, it is faster on datasets DBLP, PubMed, Cora and CS by $1.4\times$, $3.2\times$, $2.5\times$, and $2.6\times$ respectively.

% \begin{table}[htb]
% \centering
% \caption{Space efficiency (↓) of unlearning models reflected by the number of trainable parameters.}
% \label{tab:space}
% \begin{tabular}{@{}llll@{}}
% \toprule
% \multicolumn{1}{c}{Method} & \multicolumn{1}{c}{Parameters}\\ \midrule
% \rowcolor[HTML]{EFEFEF} 
% \textsc{Gold}         & $24576$ \\
% \textsc{GradAscent}   & $24576$ \\
% \textsc{D2D}          & $24576$ \\
% \textsc{GraphEraser}  & $245760$ \\
% \textsc{GraphEditor}  & $24576$ \\
% \textsc{CertUnlearn}  & $24576$ \\
% \textsc{GNNDelete}    & $256$ \\
% \name (\textit{Ours}) & $\approx 0$ \\ \bottomrule
% \end{tabular}
% \end{table}
% \textbf{Space efficiency.} We compare the number of trainable parameters a model has in Table \ref{tab:space} to highlight \name's space efficiency. \textsc{GraphEraser} requires separate models for graph shards, leading to high inefficiency. Retraining from scratch in \textsc{Gold} model typically initializes the model's parameters and trains it anew. Hence the number of parameters is the same as the original training. The parameters introduced by \textsc{GNNDelete}'s deletion layers are influenced by hidden dimensions and the number of hops, yet they don't scale directly with graph size.  \name's approach involves refining existing parameters through additional training epochs, driven by unlearning objectives and separators. Consequently, \name achieves nearly zero overhead costs by not introducing new trainable parameters.
\textbf{Inference cost.}
% \begin{figure}[t]
% \centering
% \includegraphics[width=0.35\textwidth]{fig/FLOPs.png}
% \caption{Difference in FLOPS between \name and \textsc{GNNDelete} for various datasets.}
% \label{fig:FLOPs}
% \end{figure}
We present FLOPs (floating-point operations) per forward pass, for comparing inference cost in Figure \ref{fig:FLOPs}. \textsc{GNNDelete} adds a deletion layer, increasing computational overhead. In contrast, \name uses the standard GCN architecture, avoiding extra unlearning costs. For example, in DBLP dataset, with $17$k nodes and $100$k edges, \name has about $5.20\times1e9$ FLOPs per forward pass, while \textsc{GNNDelete} has $5.21\times1e9$ FLOPs ($10.2\times1e6$ additional FLOPs). Cumulatively, this has an significant impact over multiple passes and epochs. Others like \textsc{GraphEraser}, incur even higher overhead due to partitioning and aggregation.

\begin{table}[]
\centering
\caption{\noindent AUC results for link prediction, unlearning 100 nodes on \textbf{DBLP}. Best in \textbf{bold}, second best \underline{underlined}}
\label{tab:NodeUnlearning}
% \resizebox{\linewidth}{!}{%
\begin{tabular}{@{}lc@{}}
\toprule
Model  & DBLP                                              \\ \midrule
\rowcolor[HTML]{EFEFEF}
\multicolumn{1}{l}{\textsc{Gold}}        & $0.973 \pm 0.002$                                 \\
\multicolumn{1}{l}{\textsc{Gradascent}}  & $0.571 \pm 0.032$                                 \\
\multicolumn{1}{l}{\textsc{D2d}}         & $0.507 \pm 0.002$                                 \\
\multicolumn{1}{l}{\textsc{Grapheraser}} & $0.513 \pm 0.004$                                 \\
\multicolumn{1}{l}{\textsc{Grapheditor}} & $0.697 \pm 0.031$                                 \\
\multicolumn{1}{l}{\textsc{Certunlearn}} & $0.713 \pm 0.025$                                 \\
\multicolumn{1}{l}{\textsc{GNNdelete}}   & $\mathbf{0.938 \pm 0.004}$                        \\
\multicolumn{1}{l}{\name}       & \multicolumn{1}{l}{$\underline{0.868 \pm 0.015}$} \\ \bottomrule
\end{tabular}%
\end{table}
\begin{table}[]
% \begin{minipage}{0.45\columnwidth}
% \vspace*{-\baselineskip}
% \begin{table}[H]
% \centering
% \caption{\noindent AUC results for link prediction, unlearning 100 nodes on \textbf{DBLP}. Best in \textbf{bold}, second best in \underline{underline}}
% \label{tab:NodeUnlearning}
% \vspace*{-\baselineskip}
% \resizebox{0.99\columnwidth}{!}{%
% \begin{tabular}{@{}lc@{}}
% \toprule
% Model  & DBLP                                              \\ \midrule
% \rowcolor[HTML]{EFEFEF}
% \multicolumn{1}{l}{\textsc{Gold}}        & $0.973 \pm 0.002$                                 \\
% \multicolumn{1}{l}{\textsc{Gradascent}}  & $0.571 \pm 0.032$                                 \\
% \multicolumn{1}{l}{\textsc{D2d}}         & $0.507 \pm 0.002$                                 \\
% \multicolumn{1}{l}{\textsc{Grapheraser}} & $0.513 \pm 0.004$                                 \\
% \multicolumn{1}{l}{\textsc{Grapheditor}} & $0.697 \pm 0.031$                                 \\
% \multicolumn{1}{l}{\textsc{Certunlearn}} & $0.713 \pm 0.025$                                 \\
% \multicolumn{1}{l}{\textsc{GNNdelete}}   & $\mathbf{0.938 \pm 0.004}$                        \\
% \multicolumn{1}{l}{\name}       & \multicolumn{1}{l}{$\underline{0.868 \pm 0.015}$} \\ \bottomrule
% \end{tabular}%
% }
% \end{table}
% \vspace*{-\baselineskip}
% \end{minipage}
% \hfill%
% \begin{minipage}{0.49\columnwidth}
\caption{AUC results for link prediction, unlearning up to 50\% edges on \textbf{DBLP} and \textbf{PubMed} with \name.}
% \vspace*{-\baselineskip}
\label{tab:moreedgeunlearn}
\centering
% \resizebox{0.8\columnwidth}{!}{%
\begin{tabular}{@{}clcc@{}}
\toprule
Ratio (\%)  & DBLP  & PubMed \\ \midrule
0.5\%       & 0.967 & 0.960  \\ %\midrule
2.5\%       & 0.969 & 0.958  \\ %\midrule
5\%         & 0.965 & 0.957  \\ %\midrule
10\%        & 0.940 & 0.885  \\ %\midrule
20\%        & 0.935 & 0.878  \\ %\midrule
30\%        & 0.927 & 0.871  \\ %\midrule
40\%        & 0.916 & 0.866  \\ %\midrule
50\%        & 0.900 & 0.854  \\ \bottomrule
\end{tabular}
% }
% \end{minipage}%
\end{table}
% \vspace*{-\baselineskip}
%Old table
% \begin{tabular}{@{}lccc@{}}
% \toprule 
% \multicolumn{1}{c}{Model} & AUC                           & F1                            & MI Ratio \\ \midrule
% \rowcolor[HTML]{EFEFEF} 
% \textsc{Gold}             & $0.845 \pm 0.008$             & $0.841 \pm 0.004$             & $1.515 \pm 0.034$ \\
% \textsc{GradAscent}       & $0.392 \pm 0.026$             & $0.341 \pm 0.035$             & $1.021 \pm 0.113$ \\
% \textsc{D2D}              & $0.250 \pm 0.000$             & $0.250 \pm 0.000$             & $\underline{1.755 \pm 0.065}$ \\
% \textsc{GraphEraser}      & $0.718 \pm 0.014$             & $0.716 \pm 0.011$             & $0.975 \pm 0.083$ \\
% \textsc{GraphEditor}      & $0.765 \pm 0.012$             & $0.749 \pm 0.006$             & $1.260 \pm 0.088$ \\
% \textsc{CertUnlearn}      & $0.743 \pm 0.027$             & $0.738 \pm 0.022$             & $1.134 \pm 0.009$ \\
% \textsc{GNNDelete}        & $\underline{0.793 \pm 0.016}$ & $\underline{0.768 \pm 0.009}$ & $1.401 \pm 0.082$ \\
% \name (\textit{Ours})     & $\mathbf{0.868 \pm 0.018}$    & $\mathbf{}$                   & $\mathbf{}$ \\ \bottomrule
% \end{tabular}
% }

Recently, \textsc{GIF}\cite{wu2023gif} was proposed as a new unlearning technique applicable to link prediction tasks. \name consistently outperforms \textsc{GIF} (Table \ref{tab:GIF}) on various unlearning ratios on the Cora and PubMed datasets.
\begin{table}[]
\centering
\caption{AUC results for link prediction, unlearning edges on \textbf{Cora} and \textbf{PubMed}. (Best results in \textbf{bold}).}
\label{tab:GIF}
% \vspace*{-\baselineskip}
\resizebox{\linewidth}{!}{%
% \scalebox{0.6}{
\begin{tabular}{@{}llll@{}}
\toprule
Ratio (\%)             & Model        & Cora            & PubMed          \\ \midrule
\rowcolor[HTML]{EFEFEF}
\multirow{3}{*}{0.5\%} & Gold         & $0.965 \pm 0.002$ & $0.968 \pm 0.001$ \\
                      & \name (\textit{Ours}) & $\mathbf{0.966 \pm 0.016}$ & $\mathbf{0.963 \pm 0.018}$ \\
                      & GIF          & $0.851 \pm 0.042$ & $0.837 \pm 0.063$ \\ \midrule
\rowcolor[HTML]{EFEFEF}
\multirow{3}{*}{2.5\%} & Gold         & $0.966 \pm 0.001$ & $0.967 \pm 0.001$ \\
                      & \name (\textit{Ours}) & $\mathbf{0.964 \pm 0.019}$ & $\mathbf{0.966 \pm 0.027}$ \\
                      & GIF          & $0.859 \pm 0.053$ & $0.829 \pm 0.049$ \\ \midrule
\rowcolor[HTML]{EFEFEF}
\multirow{3}{*}{5\%}   & Gold         & $0.966 \pm 0.002$ & $0.966 \pm 0.001$ \\
                      & \name (\textit{Ours}) & $\mathbf{0.965 \pm 0.021}$ & $\mathbf{0.966 \pm 0.015}$ \\
                      & GIF          & $0.863 \pm 0.035$ & $0.823 \pm 0.051$ \\ \bottomrule 
\end{tabular}
}
\end{table}

\section{Ablation Studies}
\label{sec:strategies}
\subsection{Comparison across strategies.} We compare the three strategies of \name in Table  \ref{tab:AcrossGNNsStrat} across AUC on the retain and forget sets, MI Ratio, as well as unlearning time. 
\textit{Strategies 2} and \textit{3} achieve a marginally closer AUC to the \textsc{Gold} model. Their strategy preserves the node-level features and embeddings, potentially improving the link prediction performance by preserving the structural information and local relationships. However, they may also retain more information about the deleted data, as indicated by their lower MI Ratios.
\textit{Strategy 1} is better for preventing membership inference attacks because it focuses on preserving the probabilistic distribution of the model's predictions rather than trying to match exact predictions. Unlike \textit{Strategy 2}, where two separate models, pre-trained and randomly initialized are employed, \textit{Strategy 3}'s use of a single pre-trained model for both positive and negative knowledge accelerates the unlearning process. 
\begin{table}[]
\centering
\caption{AUC results for link prediction when unlearning 2.5\% edges $\E_f = \E_{f,\mathrm{IN}}$ on DBLP dataset. The closest method to \textsc{Gold} model is marked in \textbf{bold}.}
\label{tab:AcrossGNNsStrat}
\resizebox{\linewidth}{!}{%
\begin{tabular}{@{}llllllll@{}}
\toprule
\multicolumn{1}{c}{}                        & \multicolumn{3}{c}{GCN} & \multicolumn{1}{c}{} \\ \cmidrule(lr){2-4}
\multicolumn{1}{c}{\multirow{-2}{*}{Model}} & \multicolumn{1}{c}{$\E_r$} & \multicolumn{1}{c}{$\E_f$} & \multicolumn{1}{c}{MI Ratio} & \multicolumn{1}{c}{\multirow{-2}{*}{\begin{tabular}[c]{@{}l@{}}Unlearn\\time (sec)\end{tabular}}} \\ \midrule
\rowcolor[HTML]{EFEFEF} 
\textsc{Gold} & $0.964 \pm 0.003$ & $0.506 \pm 0.013$ & $1.255 \pm 0.207$ & 5813 \\
Strategy 1 & $0.958 \pm 0.009$ & $0.501 \pm 0.103$ & $\mathbf{2.531 \pm 0.163}$ & 2573 \\
Strategy 2 & $\mathbf{0.967 \pm 0.012}$ & $0.507 \pm 0.013$ & $1.142 \pm 0.152$ & 2202 \\
Strategy 3 & $\mathbf{0.967 \pm 0.015}$ & $0.507 \pm 0.019$ & $1.138 \pm 0.108$ & 1691 \\
\bottomrule
\end{tabular}
}
\end{table}
\subsection{Scalability}
\begin{figure}[H]
  \centering
    \includegraphics[width=\linewidth]{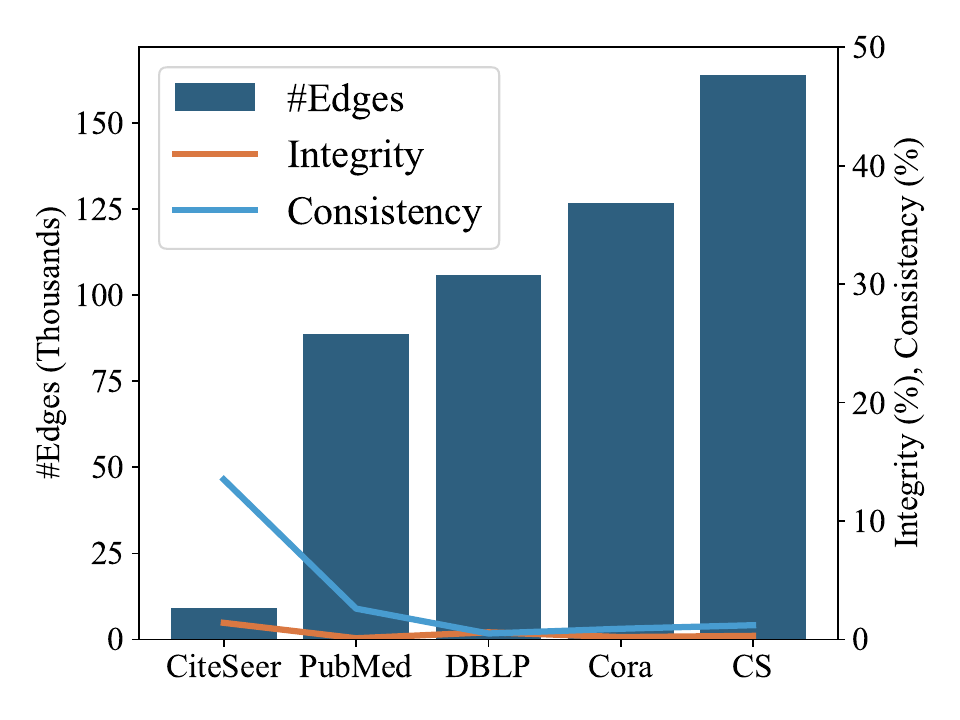}
    \caption{Integrity, Consistency as graph size increases}
    \label{fig:scale}
  \label{fig:both_pictures}
\end{figure}

Figure \ref{fig:scale} demonstrates \name's consistent effectiveness in removing the influence of forget sets and preserving elements of the retained sets, even as graph sizes increase. For both small datasets like CiteSeer with $25$ thousand edges and with large datasets like CS with $160$ thousand edges, the performance of retained set does not depart more than $2\%$ from the gold model. Hence, integrity is high. Similarly, the performance on forget set, does not depeart more than $2\%$ from the gold model, except for CiteSeer. Hence, consistency is high.
\subsection{Unlearning a higher percentage}
To assess the effects of increasing edge unlearning, we examine link prediction accuracy on the retain set. The results in Table~\ref{tab:moreedgeunlearn} demonstrate a noticeable accuracy decrease with increase in the percentage of edges unlearned, showcasing the model's adaptability to varying unlearning sizes. This enhances its versatility for real-world scenarios requiring dynamic alterations.
% \begin{table}[htb]
% \centering
% \caption{\noindent AUC results for link prediction, unlearning up to 50\% edges on \textbf{DBLP} and \textbf{PubMed} with \name.}
% \vspace*{-\baselineskip}
% \label{tab:moreedgeunlearn}
% \centering
% % \resizebox{0.9\columnwidth}{!}{%
% \begin{tabular}{@{}clcc@{}}
% \toprule
% Ratio (\%)  & DBLP  & PubMed \\ \midrule
% 0.5\%       & 0.967 & 0.960  \\ %\midrule
% 2.5\%       & 0.969 & 0.958  \\ %\midrule
% 5\%         & 0.965 & 0.957  \\ %\midrule
% 10\%        & 0.940 & 0.885  \\ %\midrule
% 20\%        & 0.935 & 0.878  \\ %\midrule
% 30\%        & 0.927 & 0.871  \\ %\midrule
% 40\%        & 0.916 & 0.866  \\ %\midrule
% 50\%        & 0.900 & 0.854  \\ \bottomrule
% \end{tabular}
% % }
% \end{table}
\subsection{Sensitivity of AUC to learning rate}
Table \ref{tab:learningRate} presents the AUC of \name using different learning rates. A high learning rate can lead to erratic updates to the model's parameters, making it difficult for the unlearned model to preserve knowledge. But such updates can move away the model from the undesired knowledge space more rapidly. Hence, the effect on $\E_f$ is not as pronounced. Thus, AUC may not be a reliable metric for evaluating unlearning in all cases. The model may still have the influences of the forgotten set, which can be detected using a more robust metric like a membership inference attack.

\begin{table}[]
\centering
\caption{Sensitivity of AUC to learning rate.}
\label{tab:learningRate}
%\resizebox{0.4\columnwidth}{!}{%
\begin{tabular}{@{}lcc@{}}
\toprule
\textbf{Learning rate} & \textbf{$\E_r$} & \textbf{$\E_f$} \\
\midrule
$0.001$ & $0.9657$ & $0.5047$ \\
$0.005$ & $0.8531$ & $0.4115$ \\
$0.01$  & $0.7177$ & $0.4194$ \\
$0.1$   & $0.7396$ & $0.4215$ \\
$1$     & $0.7204$ & $0.5165$ \\
$10$    & $0.7471$ & $0.5095$ \\
\bottomrule
\end{tabular}
%}
\end{table}
\section{Conclusions}
\label{sec:conclusion}
We present \name, a novel approach to graph unlearning using knowledge distillation. It effectively tackles the issues of poor handling of local graph dependencies and overhead costs in the existing literature. Our efficient and model-agnostic distillation method employs response-based soft targets and feature-based node embedding for distillation, while minimizing KL divergence. Through a comprehensive series of experiments on diverse benchmark datasets, we showcase that the unlearned model successfully eradicates the influence of deleted graph elements while preserving the knowledge of retained ones. \name's robustness is further confirmed by membership inference attack metrics, highlighting its ability to prevent information leakage. Further, it is remarkably faster than the best available existing method. Future work includes evaluating the performance of \name on a time-evolving dataset and unlearning in a federated learning setup.

\section*{Acknowledgment}
This research/project is supported by the National Research Foundation, Singapore under its Strategic Capability Research Centres Funding Initiative. Any opinions, findings and conclusions or recommendations expressed in this material are those of the author(s) and do not reflect the views of National Research Foundation, Singapore.

%delving into its behavior in challenging unlearning scenarios such as zero-shot unlearning.

% \input{sec/X_author_response}
\bibliographystyle{IEEEtran}
\bibliography{ijcai24}
%\pagebreak
% \appendix

% --- PDF will be split by an editor (e.g. macOS preview), so need to restart from page 1
% \setcounter{page}{1}

% --- repeat the title (AT: haven't found a more elegant way to do this...)

\twocolumn[
%\centering
%\Large
%\textbf{Supplementary Material} \\
%\vspace{0.5em} 
%\textbf{Distill to Delete: Unlearning in Graph Networks with Knowledge Distillation}
% \\
% \vspace{1.0em}
] %< twocolumn
\newpage
\appendix

% \begin{minipage}{\textwidth}
% \centering\noindent

\begin{table*}[!ht]
\caption{AUC results for link prediction when unlearning $0.5\%$, $2.5\%$ and $5.0\%$ edges $\E_f = \E_{f,\mathrm{IN}}$ on \textbf{PubMed} dataset. The closest method to \textsc{Gold} model is marked in \textbf{bold}, and the second closest is \underline{underlined}. `-' denotes that method does not apply to those GNNs.}
\label{tab:PubMedIN}
\begin{adjustbox}{width=\linewidth,center}
\begin{tabular}{@{}clcccccc@{}}
\toprule
\multirow{2}{*}{Ratio (\%)} & \multirow{2}{*}{Model} & \multicolumn{2}{c}{GCN} & \multicolumn{2}{c}{GAT} & \multicolumn{2}{c}{GIN} \\
& & $\E_r$ & $\E_f$ & $\E_r$ & $\E_f$ & $\E_r$ & $\E_f$ \\
\midrule
\rowcolor[HTML]{EFEFEF}
\multirow{7}{*}{$0.5$}
& \textsc{Gold} & $0.968 \pm 0.001$ & $0.493 \pm 0.040$ & $0.931 \pm 0.003$ & $0.533 \pm 0.037$ & $0.940 \pm 0.002$ & $0.626 \pm 0.041$ \\
& \textsc{GradAscent} & $0.469 \pm 0.095$ & $\mathbf{0.496 \pm 0.058}$ & $0.436 \pm 0.028$ & $\mathbf{0.553 \pm 0.029}$ & $0.687 \pm 0.060$ & $\underline{0.556 \pm 0.042}$ \\
& \textsc{D2D} & $0.500 \pm 0.000$ & $\underline{0.500 \pm 0.000}$ & $0.500 \pm 0.000$ & $\underline{0.500 \pm 0.000}$ & $0.500 \pm 0.000$ & $0.500 \pm 0.000$ \\
& \textsc{GraphEraser} & $0.547 \pm 0.004$ & $\underline{0.500 \pm 0.000}$ & $0.536 \pm 0.000$ & $\underline{0.500 \pm 0.000}$ & $0.524 \pm 0.002$ & $0.500 \pm 0.000$ \\
& \textsc{GraphEditor} & $0.669 \pm 0.005$ & $0.469 \pm 0.021$ & - & - & - & - \\
& \textsc{CertUnlearn} & $0.657 \pm 0.015$ & $0.515 \pm 0.027$ & - & - & - & - \\
& \textsc{GNNDelete} & $\underline{0.951 \pm 0.005}$ & $0.838 \pm 0.014$ & $\underline{0.909 \pm 0.003}$ & $0.888 \pm 0.016$ & $\underline{0.929 \pm 0.006}$ & $0.835 \pm 0.006$ \\
& \name (\textit{Ours}) & $\mathbf{0.967 \pm 0.015}$ & $0.552 \pm 0.033$ & $\mathbf{0.933 \pm 0.029}$ & $0.610 \pm 0.015$ & $\mathbf{0.940 \pm 0.024}$ & $\mathbf{0.668 \pm 0.012}$ \\
\midrule
\rowcolor[HTML]{EFEFEF}
\multirow{7}{*}{$2.5$}
& \textsc{Gold} & $0.968 \pm 0.001$ & $0.499 \pm 0.019$ & $0.931 \pm 0.002$ & $0.541 \pm 0.013$ & $0.937 \pm 0.004$ & $0.614 \pm 0.015$ \\
& \textsc{GradAscent} & $0.470 \pm 0.087$ & $\underline{0.474 \pm 0.039}$ & $0.522 \pm 0.066$ & $0.704 \pm 0.086$ & $0.631 \pm 0.050$ & $0.499 \pm 0.018$ \\
& \textsc{D2D} & $0.500 \pm 0.000$ & $\mathbf{0.500 \pm 0.000}$ & $0.500 \pm 0.000$ & $\underline{0.500 \pm 0.000}$ & $0.500 \pm 0.000$ & $\underline{0.500 \pm 0.000}$ \\
& \textsc{GraphEraser} & $0.538 \pm 0.003$ & $\mathbf{0.500 \pm 0.000}$ & $0.521 \pm 0.003$ & $\underline{0.500 \pm 0.000}$ & $0.533 \pm 0.010$ & $\underline{0.500 \pm 0.000}$ \\
& \textsc{GraphEditor} & $0.657 \pm 0.006$ & $0.467 \pm 0.006$ & - & - & - & - \\
& \textsc{CertUnlearn} & $0.622 \pm 0.009$ & $0.468 \pm 0.025$ & - & - & - & - \\
& \textsc{GNNDelete} & $0.920 \pm 0.014$ & $0.739 \pm 0.010$ & $\underline{0.891 \pm 0.005}$ & $0.759 \pm 0.012$ & $\underline{0.909 \pm 0.005}$ & $0.782 \pm 0.013$ \\
& \name (\textit{Ours}) & $\mathbf{0.969 \pm 0.019}$ & $0.525 \pm 0.053$ & $\mathbf{0.932 \pm 0.018}$ & $\mathbf{0.575 \pm 0.016}$ & $\mathbf{0.941 \pm 0.021}$ & $\mathbf{0.607 \pm 0.018}$ \\
\midrule
\rowcolor[HTML]{EFEFEF}
\multirow{7}{*}{$5.0$}
& \textsc{Gold} & $0.967 \pm 0.001$ & $0.503 \pm 0.009$ & $0.929 \pm 0.003$ & $0.545 \pm 0.005$ & $0.936 \pm 0.005$ & $0.621 \pm 0.003$ \\
& \textsc{GradAscent} & $0.473 \pm 0.090$ & $0.473 \pm 0.038$ & $0.525 \pm 0.069$ & $\underline{0.686 \pm 0.090}$ & $0.635 \pm 0.073$ & $0.493 \pm 0.018$ \\
& \textsc{D2D} & $0.500 \pm 0.000$ & $\mathbf{0.500 \pm 0.000}$ & $0.500 \pm 0.000$ & $0.500 \pm 0.000$ & $0.500 \pm 0.000$ & $\underline{0.500 \pm 0.000}$ \\
& \textsc{GraphEraser} & $0.551 \pm 0.004$ & $\mathbf{0.500 \pm 0.000}$ & $0.524 \pm 0.020$ & $0.500 \pm 0.000$ & $0.531 \pm 0.000$ & $\underline{0.500 \pm 0.000}$ \\
& \textsc{GraphEditor} & $0.556 \pm 0.007$ & $0.468 \pm 0.002$ & - & - & - & - \\
& \textsc{CertUnlearn} & $0.572 \pm 0.013$ & $0.477 \pm 0.028$ & - & - & - & - \\
& \textsc{GNNDelete} & $\underline{0.916 \pm 0.006}$ & $0.691 \pm 0.012$ & $\underline{0.887 \pm 0.009}$ & $0.713 \pm 0.005$ & $\underline{0.895 \pm 0.004}$ & $0.761 \pm 0.005$ \\
& \name (\textit{Ours}) & $\mathbf{0.965 \pm 0.008}$ & $\underline{0.511 \pm 0.043}$ & $\mathbf{0.917 \pm 0.011}$ & $\mathbf{0.574 \pm 0.019}$ & $\mathbf{0.941 \pm 0.013}$ & $\mathbf{0.603 \pm 0.017}$ \\
\bottomrule
\end{tabular}
\end{adjustbox}
% \end{table*}
\bigskip
% \begin{table*}[htb]
\caption{AUC results for link prediction when unlearning $0.5\%$, $2.5\%$ and $5.0\%$ edges $\E_f = \E_{f,\mathrm{OUT}}$ on \textbf{PubMed} dataset. The closest method to \textsc{Gold} model is marked in \textbf{bold}, and the second closest is \underline{underlined}. `-' denotes that the method does not apply to those GNNs.}
\label{tab:PubMedOUT}
\resizebox{\linewidth}{!}{%
\begin{tabular}{@{}clcccccc@{}}
\toprule
\multirow{2}{*}{Ratio (\%)} & \multirow{2}{*}{Model} & \multicolumn{2}{c}{GCN} & \multicolumn{2}{c}{GAT} & \multicolumn{2}{c}{GIN} \\
 & & $\E_r$ & $\E_f$ & $\E_r$ & $\E_f$ & $\E_r$ & $\E_f$ \\
\midrule
\rowcolor[HTML]{EFEFEF}
\multirow{7}{*}{$0.5$}
 & \textsc{Gold} & $0.968 \pm 0.001$ & $0.687 \pm 0.023$ & $0.931 \pm 0.003$ & $0.723 \pm 0.026$ & $0.941 \pm 0.004$ & $0.865 \pm 0.012$ \\
 & \textsc{GradAscent} & $0.458 \pm 0.139$ & $0.539 \pm 0.091$ & $0.450 \pm 0.017$ & $\underline{0.541 \pm 0.049}$ & $0.518 \pm 0.122$ & $0.528 \pm 0.021$ \\
 & \textsc{D2D} & $0.500 \pm 0.000$ & $0.500 \pm 0.000$ & $0.500 \pm 0.000$ & $0.500 \pm 0.000$ & $0.500 \pm 0.000$ & $0.500 \pm 0.000$ \\
 & \textsc{GraphEraser} & $0.529 \pm 0.013$ & $0.500 \pm 0.000$ & $0.542 \pm 0.004$ & $0.500 \pm 0.000$ & $\underline{0.535 \pm 0.003}$ & $0.500 \pm 0.000$ \\
 & \textsc{GraphEditor} & $0.732 \pm 0.043$ & $0.603 \pm 0.015$ & - & - & - & - \\
 & \textsc{CertUnlearn} & $0.724 \pm 0.012$ & $\underline{0.597 \pm 0.029}$ & - & - & - & - \\
 & \textsc{GNNDelete} & $\underline{0.961 \pm 0.004}$ & $0.973 \pm 0.005$ & $\underline{0.926 \pm 0.006}$ & $0.976 \pm 0.005$ & $\mathbf{0.940 \pm 0.005}$ & $\underline{0.963 \pm 0.010}$ \\
 & \name (\textit{Ours}) & $\mathbf{0.963 \pm 0.018}$ & $\mathbf{0.685 \pm 0.013}$ & $\mathbf{0.933 \pm 0.015}$ & $\mathbf{0.706 \pm 0.034}$ & $\mathbf{0.940 \pm 0.024}$ & $\mathbf{0.836 \pm 0.009}$ \\
\midrule
\rowcolor[HTML]{EFEFEF}
\multirow{7}{*}{$2.5$}
 & \textsc{Gold} & $0.967 \pm 0.001$ & $0.696 \pm 0.011$ & $0.930 \pm 0.003$ & $0.736 \pm 0.011$ & $0.942 \pm 0.005$ & $0.875 \pm 0.008$ \\
 & \textsc{GradAscent} & $0.446 \pm 0.130$ & $0.500 \pm 0.067$ & $0.582 \pm 0.006$ & $\underline{0.758 \pm 0.033}$ & $0.406 \pm 0.054$ & $0.454 \pm 0.037$ \\
 & \textsc{D2D} & $0.500 \pm 0.000$ & $0.500 \pm 0.000$ & $0.500 \pm 0.000$ & $0.500 \pm 0.000$ & $0.500 \pm 0.000$ & $0.500 \pm 0.000$ \\
 & \textsc{GraphEraser} & $0.505 \pm 0.024$ & $0.500 \pm 0.000$ & $0.538 \pm 0.009$ & $0.500 \pm 0.000$ & $\underline{0.544 \pm 0.014}$ & $0.500 \pm 0.000$ \\
 & \textsc{GraphEditor} & $0.689 \pm 0.015$ & $0.570 \pm 0.011$ & - & - & - & - \\
 & \textsc{CertUnlearn} & $0.697 \pm 0.012$ & $\underline{0.582 \pm 0.032}$ & - & - & - & - \\
 & \textsc{GNNDelete} & $\underline{0.954 \pm 0.003}$ & $0.909 \pm 0.004$ & $\underline{0.920 \pm 0.004}$ & $0.916 \pm 0.006$ & $\mathbf{0.943 \pm 0.005}$ & $\underline{0.938 \pm 0.009}$ \\
 & \name (\textit{Ours}) & $\mathbf{0.966 \pm 0.027}$ & $\mathbf{0.662 \pm 0.021}$ & $\mathbf{0.932 \pm 0.033}$ & $\mathbf{0.715 \pm 0.013}$ & $\mathbf{0.941 \pm 0.022}$ & $\mathbf{0.852 \pm 0.021}$ \\
\midrule
\rowcolor[HTML]{EFEFEF}
\multirow{7}{*}{$5.0$}
 & \textsc{Gold} & $0.966 \pm 0.001$ & $0.707 \pm 0.004$ & $0.929 \pm 0.002$ & $0.744 \pm 0.008$ & $0.942 \pm 0.004$ & $0.885 \pm 0.010$ \\
 & \textsc{GradAscent} & $0.446 \pm 0.126$ & $0.492 \pm 0.064$ & $0.581 \pm 0.010$ & $\underline{0.704 \pm 0.022}$ & $0.388 \pm 0.056$ & $0.455 \pm 0.028$ \\
 & \textsc{D2D} & $0.500 \pm 0.000$ & $0.500 \pm 0.000$ & $0.500 \pm 0.000$ & $0.500 \pm 0.000$ & $0.500 \pm 0.000$ & $0.500 \pm 0.000$ \\
 & \textsc{GraphEraser} & $0.532 \pm 0.001$ & $0.500 \pm 0.000$ & $0.527 \pm 0.022$ & $0.500 \pm 0.000$ & $0.524 \pm 0.015$ & $0.500 \pm 0.000$ \\
 & \textsc{GraphEditor} & $0.598 \pm 0.023$ & $0.530 \pm 0.006$ & - & - & - & - \\
 & \textsc{CertUnlearn} & $0.643 \pm 0.031$ & $0.534 \pm 0.020$ & - & - & - & - \\
 & \textsc{GNNDelete} & $\underline{0.950 \pm 0.003}$ & $\underline{0.859 \pm 0.005}$ & $\underline{0.921 \pm 0.005}$ & $0.863 \pm 0.006$ & $\underline{0.941 \pm 0.002}$ & $\underline{0.930 \pm 0.009}$ \\
 & \name (\textit{Ours}) & $\mathbf{0.966 \pm 0.015}$ & $\mathbf{0.664 \pm 0.021}$ & $\mathbf{0.931 \pm 0.017}$ & $\mathbf{0.715 \pm 0.024}$ & $\mathbf{0.942 \pm 0.023}$ & $\mathbf{0.864 \pm 0.015}$ \\
\bottomrule
\end{tabular}
}
\end{table*}
% \bigskip

\begin{table*}[!ht]
\caption{AUC results for link prediction when unlearning 0.5\%, 2.5\%, and 5.0\% edges $\mathcal{E}_f = \mathcal{E}_{f,\mathrm{IN}}$ on the \textbf{Cora} dataset. The closest method to the \textsc{Gold} model is marked in \textbf{bold}, and the second closest is \underline{underlined}. `-' denotes that the method does not apply to those GNNs.}
\label{tab:CoraIN}
\resizebox{\linewidth}{!}{%
\begin{tabular}{@{}clcccccc@{}}
\toprule
\multirow{2}{*}{$\begin{array}{l}\text{Ratio}\\(\%)\end{array}$} & \multirow{2}{*}{Model} & \multicolumn{2}{c}{GCN} & \multicolumn{2}{c}{GAT} & \multicolumn{2}{c}{GIN} \\
 & & $\mathcal{E}_r$ & $\mathcal{E}_f$ & $\mathcal{E}_r$ & $\mathcal{E}_f$ & $\mathcal{E}_r$ & $\mathcal{E}_f$ \\
\midrule
\rowcolor[HTML]{EFEFEF}
\multirow{7}{*}{$0.5$} 
 & \textsc{Gold} & $0.965 \pm 0.002$ & $0.511 \pm 0.024$ & $0.961 \pm 0.001$ & $0.513 \pm 0.024$ & $0.960 \pm 0.003$ & $0.571 \pm 0.028$ \\
 & \textsc{GradAscent} & $0.528 \pm 0.008$ & $0.588 \pm 0.014$ & $0.502 \pm 0.002$ & $\underline{0.543 \pm 0.058}$ & $0.792 \pm 0.046$ & $0.705 \pm 0.113$ \\
 & \textsc{D2D} & $0.500 \pm 0.000$ & $0.500 \pm 0.000$ & $0.500 \pm 0.000$ & $\mathbf{0.500 \pm 0.000}$ & $0.500 \pm 0.000$ & $\underline{0.500 \pm 0.000}$ \\
 & \textsc{GraphEraser} & $0.528 \pm 0.002$ & $0.500 \pm 0.000$ & $0.523 \pm 0.013$ & $\mathbf{0.500 \pm 0.000}$ & $0.542 \pm 0.009$ & $\underline{0.500 \pm 0.000}$ \\
 & \textsc{GraphEditor} & $0.704 \pm 0.057$ & $0.488 \pm 0.024$ & - & - & - & - \\
 & \textsc{CertUnlearn} & $0.811 \pm 0.035$ & $\underline{0.497 \pm 0.013}$ & - & - & - & - \\
 & \textsc{GNNDelete} & $\underline{0.944 \pm 0.003}$ & $0.843 \pm 0.015$ & $\underline{0.937 \pm 0.004}$ & $0.880 \pm 0.011$ & $\underline{0.942 \pm 0.005}$ & $0.824 \pm 0.021$ \\
 & \name \textit{(Ours)} & $\mathbf{0.966 \pm 0.023}$ & $\mathbf{0.510 \pm 0.017}$ & $\mathbf{0.963 \pm 0.013}$ & $0.496 \pm 0.023$ & $\mathbf{0.956 \pm 0.011}$ & $\mathbf{0.590 \pm 0.019}$ \\
\midrule
\rowcolor[HTML]{EFEFEF}
\multirow{7}{*}{$2.5$} 
 & \textsc{Gold} & $0.966 \pm 0.002$ & $0.520 \pm 0.008$ & $0.961 \pm 0.001$ & $0.520 \pm 0.012$ & $0.958 \pm 0.002$ & $0.583 \pm 0.007$ \\
 & \textsc{GradAscent} & $0.509 \pm 0.006$ & $\underline{0.509 \pm 0.007}$ & $0.490 \pm 0.007$ & $\underline{0.551 \pm 0.014}$ & $0.639 \pm 0.077$ & $\underline{0.614 \pm 0.016}$ \\
 & \textsc{D2D} & $0.500 \pm 0.000$ & $0.500 \pm 0.000$ & $0.500 \pm 0.000$ & $\mathbf{0.500 \pm 0.000}$ & $0.500 \pm 0.000$ & $0.500 \pm 0.000$ \\
 & \textsc{GraphEraser} & $0.517 \pm 0.002$ & $0.500 \pm 0.000$ & $0.556 \pm 0.013$ & $\mathbf{0.500 \pm 0.000}$ & $0.547 \pm 0.009$ & $0.500 \pm 0.000$ \\
 & \textsc{GraphEditor} & $0.673 \pm 0.091$ & $0.493 \pm 0.027$ & - & - & - & - \\
 & \textsc{CertUnlearn} & $0.781 \pm 0.042$ & $0.492 \pm 0.015$ & - & - & - & - \\
 & \textsc{GNNDelete} & $\underline{0.925 \pm 0.006}$ & $0.716 \pm 0.003$ & $\underline{0.928 \pm 0.007}$ & $0.738 \pm 0.005$ & $\underline{0.919 \pm 0.004}$ & $0.745 \pm 0.005$ \\
 & \name \textit{(Ours)} & $\mathbf{0.964 \pm 0.025}$ & $\mathbf{0.511 \pm 0.023}$ & $\mathbf{0.963 \pm 0.011}$ & $0.498 \pm 0.015$ & $\mathbf{0.957 \pm 0.044}$ & $\mathbf{0.590 \pm 0.018}$ \\
\midrule
\rowcolor[HTML]{EFEFEF}
\multirow{7}{*}{$5.0$} 
 & \textsc{Gold} & $0.964 \pm 0.002$ & $0.525 \pm 0.008$ & $0.960 \pm 0.001$ & $0.525 \pm 0.007$ & $0.958 \pm 0.002$ & $0.591 \pm 0.006$ \\
 & \textsc{GradAscent} & $0.509 \pm 0.005$ & $0.487 \pm 0.003$ & $0.489 \pm 0.015$ & $\mathbf{0.537 \pm 0.007}$ & $0.592 \pm 0.031$ & $\mathbf{0.583 \pm 0.013}$ \\
 & \textsc{D2D} & $0.500 \pm 0.000$ & $\underline{0.500 \pm 0.000}$ & $0.500 \pm 0.000$ & $\underline{0.500 \pm 0.000}$ & $0.500 \pm 0.000$ & $0.500 \pm 0.000$ \\
 & \textsc{GraphEraser} & $0.528 \pm 0.002$ & $\underline{0.500 \pm 0.000}$ & $0.517 \pm 0.013$ & $\underline{0.500 \pm 0.000}$ & $0.530 \pm 0.009$ & $0.500 \pm 0.000$ \\
 & \textsc{GraphEditor} & $0.587 \pm 0.014$ & $0.475 \pm 0.015$ & - & - & - & - \\
 & \textsc{CertUnlearn} & $0.664 \pm 0.023$ & $0.457 \pm 0.021$ & - & - & - & - \\
 & \textsc{GNNDelete} & $\underline{0.916 \pm 0.007}$ & $0.680 \pm 0.006$ & $\underline{0.920 \pm 0.005}$ & $0.700 \pm 0.004$ & $\underline{0.900 \pm 0.005}$ & $0.717 \pm 0.003$ \\
 & \name \textit{(Ours)} & $\mathbf{0.965 \pm 0.011}$ & $\mathbf{0.510 \pm 0.019}$ & $\mathbf{0.962 \pm 0.023}$ & $0.492 \pm 0.013$ & $\mathbf{0.956 \pm 0.032}$ & $\underline{0.602 \pm 0.012}$ \\
\bottomrule
\end{tabular}
}
% \end{table*}
\bigskip

% \begin{table*}[htb]
\caption{AUC results for link prediction when unlearning 0.5\%, 2.5\%, and 5.0\% edges $\mathcal{E}_f = \mathcal{E}_{f,\mathrm{OUT}}$ on the \textbf{Cora} dataset. The closest method to the \textsc{Gold} model is marked in \textbf{bold}, and the second closest is \underline{underlined}. `-' denotes that the method does not apply to those GNNs.}
\label{tab:CoraOUT}
\resizebox{\linewidth}{!}{%
\begin{tabular}{@{}clcccccc@{}}
\toprule
\multirow{2}{*}{$\begin{array}{l}\text{Ratio}\\(\%)\end{array}$} & \multirow{2}{*}{Model} & \multicolumn{2}{c}{GCN}       & \multicolumn{2}{c}{GAT}       & \multicolumn{2}{c}{GIN} \\
 &                        & $\mathcal{E}_r$               & $\mathcal{E}_f$               & $\mathcal{E}_r$               & $\mathcal{E}_f$                   & $\mathcal{E}_r$               & $\mathcal{E}_f$ \\
\midrule
\rowcolor[HTML]{EFEFEF}
\multirow{7}{*}{$0.5$} 
 & \textsc{Gold}          & $0.965 \pm 0.002$             & $0.783 \pm 0.018$             & $0.961 \pm 0.002$             & $0.756 \pm 0.013$                 & $0.961 \pm 0.002$             & $0.815 \pm 0.015$ \\
 & \textsc{GradAscent}    & $0.536 \pm 0.010$             & $\underline{0.618 \pm 0.014}$ & $0.517 \pm 0.017$             & $\underline{0.558 \pm 0.034}$     & $0.751 \pm 0.049$             & $\underline{0.778 \pm 0.043}$ \\
 & \textsc{D2D}           & $0.500 \pm 0.000$             & $0.500 \pm 0.000$             & $0.500 \pm 0.000$             & $0.500 \pm 0.000$                 & $0.500 \pm 0.000$             & $0.500 \pm 0.000$ \\
 & \textsc{GraphEraser}   & $0.563 \pm 0.002$             & $0.500 \pm 0.000$             & $0.553 \pm 0.013$             & $0.500 \pm 0.000$                 & $0.554 \pm 0.009$             & $0.500 \pm 0.000$ \\
 & \textsc{GraphEditor}   & $0.805 \pm 0.077$             & $0.614 \pm 0.054$             & -                             & -                                 & -                             & - \\
 & \textsc{CertUnlearn}   & $0.814 \pm 0.065$             & $0.603 \pm 0.039$             & -                             & -                                 & -                             & - \\
 & \textsc{GNNDelete}     & $\underline{0.958 \pm 0.002}$ & $0.977 \pm 0.001$             & $\underline{0.953 \pm 0.002}$ & $0.979 \pm 0.001$                 & $\underline{0.956 \pm 0.003}$ & $0.953 \pm 0.010$ \\
 & \name \textit{(Ours)}  & $\mathbf{0.966 \pm 0.016}$    & $\mathbf{0.737 \pm 0.019}$    & $\mathbf{0.962 \pm 0.021}$    & $\mathbf{0.721 \pm 0.015}$        & $\mathbf{0.958 \pm 0.013}$    & $\mathbf{0.820 \pm 0.017}$ \\
\midrule
\rowcolor[HTML]{EFEFEF}
\multirow{7}{*}{$2.5$} 
 & \textsc{Gold}          & $0.966 \pm 0.001$             & $0.790 \pm 0.009$             & $0.961 \pm 0.002$             & $0.758 \pm 0.011$                 & $0.961 \pm 0.003$             & $0.833 \pm 0.010$ \\
 & \textsc{GradAscent}    & $0.504 \pm 0.002$             & $0.494 \pm 0.004$             & $0.510 \pm 0.019$             & $0.522 \pm 0.023$                 & $0.603 \pm 0.039$             & $0.605 \pm 0.030$ \\
 & \textsc{D2D}           & $0.500 \pm 0.000$             & $0.500 \pm 0.000$             & $0.500 \pm 0.000$             & $0.500 \pm 0.000$                 & $0.500 \pm 0.000$             & $0.500 \pm 0.000$ \\
 & \textsc{GraphEraser}   & $0.542 \pm 0.002$             & $0.500 \pm 0.000$             & $0.519 \pm 0.013$             & $0.500 \pm 0.000$                 & $0.563 \pm 0.009$             & $0.500 \pm 0.000$ \\
 & \textsc{GraphEditor}   & $0.754 \pm 0.023$             & $0.583 \pm 0.056$             & -                             & -                                 & -                             & - \\
 & \textsc{CertUnlearn}   & $0.795 \pm 0.037$             & $0.578 \pm 0.015$             & -                             & -                                 & -                             & - \\
 & \textsc{GNNDelete}     & $\underline{0.953 \pm 0.002}$ & $\underline{0.912 \pm 0.004}$ & $\underline{0.949 \pm 0.003}$ & $\underline{0.914 \pm 0 .} 0 0 4$ & $\underline{0.953 \pm 0.002}$ & $\underline{0.922 \pm 0.006}$ \\
 & \name \textit{(Ours)}  & $\mathbf{0.964 \pm 0.019}$    & $\mathbf{0.755 \pm 0.011}$    & $\mathbf{0.963 \pm 0.029}$    & $\mathbf{0.727 \pm 0.027}$        & $\mathbf{0.958 \pm 0.018}$    & $\mathbf{0.847 \pm 0.023}$ \\
\midrule
\rowcolor[HTML]{EFEFEF}
\multirow{7}{*}{$5.0$} 
 & \textsc{Gold}          & $0.966 \pm 0.002$             & $0.812 \pm 0.006$             & $0.961 \pm 0.001$             & $0.778 \pm 0.006$                 & $0.960 \pm 0.003$             & $0.852 \pm 0.006$ \\
 & \textsc{GradAscent}    & $0.557 \pm 0.122$             & $0.513 \pm 0.106$             & $0.520 \pm 0.042$             & $0.517 \pm 0.036$                 & $0.580 \pm 0.027$             & $0.572 \pm 0.018$ \\
 & \textsc{D2D}           & $0.500 \pm 0.000$             & $0.500 \pm 0.000$             & $0.500 \pm 0.000$             & $0.500 \pm 0.000$                 & $0.500 \pm 0.000$             & $0.500 \pm 0.000$ \\
 & \textsc{GraphEraser}   & $0.514 \pm 0.002$             & $0.500 \pm 0.000$             & $\mathbf{0.523 \pm 0.013}$    & $0.500 \pm 0.000$                 & $0.533 \pm 0.009$             & $0.500 \pm 0.000$ \\
 & \textsc{GraphEditor}   & $0.721 \pm 0.048$             & $0.545 \pm 0.056$             & -                             & -                                 & -                             & - \\
 & \textsc{CertUnlearn}   & $0.745 \pm 0.033$             & $0.513 \pm 0.012$             & -                             & -                                 & -                             & - \\
 & \textsc{GNNDelete}     & $\underline{0.953 \pm 0.003}$ & $\underline{0.882 \pm 0.005}$ & $\underline{0.951 \pm 0.002}$ & $\underline{0.872 \pm 0.004}$     & $\underline{0.950 \pm 0.003}$ & $\underline{0.914 \pm 0.004}$ \\
 & \name \textit{(Ours)}  & $\mathbf{0.965 \pm 0.021}$    & $\mathbf{0.767 \pm 0.022}$    & $\mathbf{0.962 \pm 0.023}$    & $\mathbf{0.738 \pm 0.024}$        & $\mathbf{0.959 \pm 0.022}$    & $\mathbf{0.869 \pm 0.018}$ \\
\bottomrule
\end{tabular}
}
\end{table*}

\begin{table*}[!ht]
\caption{AUC results for link prediction when unlearning 0.5\%, 2.5\% and 5.0\% edges $\E_f = \E_{f,\mathrm{IN}}$ on \textbf{DBLP} dataset. The closest method to \textsc{Gold} model is marked in \textbf{bold}, and the second closest is \underline{underlined}. `-' denotes that method does not apply to those GNNs.}
\label{tab:DBLPIN}
\resizebox{\linewidth}{!}{%
\begin{tabular}{@{}clcccccc@{}}
\toprule
\multirow{2}{*}{$\begin{array}{l}\text{Ratio}\\(\%)\end{array}$} & \multirow{2}{*}{ Model } & \multicolumn{2}{c}{ GCN }     & \multicolumn{2}{c}{ GAT }     & \multicolumn{2}{c}{ GIN } \\
 &                          & $\E_r$                        & $\E_f$                        & $\E_r$                        & $\E_f$                        & $\E_r$                        & $\E_f$ \\
\midrule
\rowcolor[HTML]{EFEFEF}
\multirow{7}{*}{$0.5$} 
 & \textsc{Gold}            & $0.965 \pm 0.003$             & $0.496 \pm 0.028$             & $0.957 \pm 0.002$             & $0.513 \pm 0.021$             & $0.934 \pm 0.005$             & $0.571 \pm 0.035$ \\
 & \textsc{GradAscent}      & $0.556 \pm 0.018$             & $0.657 \pm 0.008$             & $0.511 \pm 0.023$             & $0.612 \pm 0.107$             & $0.678 \pm 0.084$             & $\mathbf{0.573 \pm 0.045}$ \\
 & \textsc{D2D}             & $0.500 \pm 0.000$             & $0.500 \pm 0.000$             & $0.500 \pm 0.000$             & $\underline{0.500 \pm 0.000}$ & $0.500 \pm 0.000$             & $0.500 \pm 0.000$ \\
 & \textsc{GraphEraser}     & $0.515 \pm 0.001$             & $0.500 \pm 0.000$             & $0.523 \pm 0.000$             & $\underline{0.500 \pm 0.000}$ & $0.507 \pm 0.003$             & $0.500 \pm 0.000$ \\
 & \textsc{GraphEditor}     & $0.781 \pm 0.026$             & $\underline{0.479 \pm 0.017}$ & -                             & -                             & -                             & - \\
 & \textsc{CertUnlearn}     & $0.742 \pm 0.021$             & $\mathbf{0.482 \pm 0.013}$    & -                             & -                             & -                             & - \\
 & \textsc{GNNDelete}       & $\underline{0.951 \pm 0.002}$ & $0.829 \pm 0.006$             & $\underline{0.928 \pm 0.004}$ & $0.889 \pm 0.011$             & $\underline{0.906 \pm 0.009}$ & $0.736 \pm 0.012$ \\
 & \name (\textit{Ours})    & $\mathbf{0.960 \pm 0.023}$    & $\underline{0.479 \pm 0.083}$ & $\mathbf{0.935 \pm 0.011}$    & $\mathbf{0.541 \pm 0.023}$    & $\mathbf{0.938 \pm 0.025}$    & $\underline{0.534 \pm 0.015}$ \\
\midrule
\rowcolor[HTML]{EFEFEF}
\multirow{7}{*}{$2.5$} 
 & \textsc{Gold}            & $0.964 \pm 0.003$             & $0.506 \pm 0.013$             & $0.956 \pm 0.002$             & $0.525 \pm 0.012$             & $0.931 \pm 0.005$             & $0.581 \pm 0.014$ \\
 & \textsc{GradAscent}      & $0.555 \pm 0.066$             & $0.594 \pm 0.063$             & $0.501 \pm 0.020$             & $0.592 \pm 0.017$             & $0.700 \pm 0.025$             & $\underline{0.524 \pm 0.017}$ \\
 & \textsc{D2D}             & $0.500 \pm 0.000$             & $\underline{0.500 \pm 0.000}$ & $0.500 \pm 0.000$             & $\underline{0.500 \pm 0.000}$ & $0.500 \pm 0.000$             & $\underline{0.524 \pm 0.017}$ \\
 & \textsc{GraphEraser}     & $0.527 \pm 0.002$             & $\underline{0.500 \pm 0.000}$ & $0.538 \pm 0.013$             & $\underline{0.500 \pm 0.000}$ & $0.517 \pm 0.009$             & $0.500 \pm 0.000$ \\
 & \textsc{GraphEditor}     & $0.776 \pm 0.025$             & $0.432 \pm 0.009$             & -                             & -                             & -                             & - \\
 & \textsc{CertUnlearn}     & $0.718 \pm 0.032$             & $0.475 \pm 0.011$             & -                             & -                             & -                             & - \\
 & \textsc{GNNDelete}       & $\underline{0.934 \pm 0.002}$ & $0.748 \pm 0.006$             & $\underline{0.914 \pm 0.007}$ & $0.774 \pm 0.015$             & $\underline{0.897 \pm 0.006}$ & $0.740 \pm 0.015$ \\
 & \name (\textit{Ours})    & $\mathbf{0.958 \pm 0.009}$    & $\mathbf{0.501 \pm 0.103}$    & $\mathbf{0.939 \pm 0.019}$    & $\mathbf{0.534 \pm 0.010}$    & $\mathbf{0.937 \pm 0.014}$    & $\mathbf{0.541 \pm 0.011}$ \\
\midrule
\rowcolor[HTML]{EFEFEF}
\multirow{7}{*}{$5.0$} 
 & \textsc{Gold}            & $0.963 \pm 0.003$             & $0.504 \pm 0.006$             & $0.955 \pm 0.002$             & $0.528 \pm 0.007$             & $0.931 \pm 0.006$             & $0.578 \pm 0.009$ \\
 & \textsc{GradAscent}      & $0.555 \pm 0.060$             & $0.581 \pm 0.073$             & $0.490 \pm 0.022$             & $\underline{0.551 \pm 0.030}$ & $0.723 \pm 0.032$             & $\underline{0.516 \pm 0.042}$ \\
 & \textsc{D2D}             & $0.500 \pm 0.000$             & $\mathbf{0.500 \pm 0.000}$    & $0.500 \pm 0.000$             & $0.500 \pm 0.000$             & $0.500 \pm 0.000$             & $0.500 \pm 0.000$ \\
 & \textsc{GraphEraser}     & $0.509 \pm 0.011$             & $\mathbf{0.500 \pm 0.000}$    & $0.511 \pm 0.006$             & $0.500 \pm 0.000$             & $0.503 \pm 0.000$             & $0.500 \pm 0.000$ \\
 & \textsc{GraphEditor}     & $0.736 \pm 0.023$             & $0.430 \pm 0.011$             & -                             & -                             & -                             & - \\
 & \textsc{CertUnlearn}     & $0.694 \pm 0.026$             & $0.441 \pm 0.008$             & -                             & -                             & -                             & - \\
 & \textsc{GNNDelete}       & $\underline{0.917 \pm 0.005}$ & $0.713 \pm 0.007$             & $\underline{0.912 \pm 0.007}$ & $0.733 \pm 0.018$             & $\underline{0.864 \pm 0.005}$ & $0.732 \pm 0.008$ \\
 & \name (\textit{Ours})    & $\mathbf{0.957 \pm 0.021}$    & $\underline{0.497 \pm 0.098}$ & $\mathbf{0.941 \pm 0.035}$    & $\mathbf{0.528 \pm 0.053}$    & $\mathbf{0.935 \pm 0.017}$    & $\mathbf{0.551 \pm 0.018}$ \\
\bottomrule
\end{tabular}
}
% \end{table*}
\bigskip
% \begin{table*}[htb]
\caption{AUC results for link prediction when unlearning 0.5\%, 2.5\% and 5.0\% edges $\E_f = \E_{f,\mathrm{OUT}}$ on \textbf{DBLP} dataset. The closest method to \textsc{Gold} model is marked in \textbf{bold}, and the second closest is \underline{underlined}. `-' denotes that method does not apply to those GNNs.}
\label{tab:DBLPOUT}
\resizebox{\linewidth}{!}{%
\begin{tabular}{@{}clcccccc@{}}
\toprule
\multirow{2}{*}{$\begin{array}{l}\text{Ratio}\\(\%)\end{array}$} & \multirow{2}{*}{ Model } & \multicolumn{2}{c}{ GCN }     & \multicolumn{2}{c}{ GAT }     & \multicolumn{2}{c}{ GIN } \\
 &                          & $\E_r$                        & $\E_f$                        & $\E_r$                        & $\E_f$                        & $\E_r$                        & $\E_f$ \\
\midrule
\rowcolor[HTML]{EFEFEF}
\multirow{7}{*}{$0.5$} 
 & \textsc{Gold}            & $0.965 \pm 0.002$             & $0.783 \pm 0.018$             & $0.956 \pm 0.002$             & $0.744 \pm 0.021$             & $0.934 \pm 0.003$             & $0.861 \pm 0.019$ \\
 & \textsc{GradAscent}      & $0.567 \pm 0.008$             & $\underline{0.696 \pm 0.017}$ & $0.501 \pm 0.030$             & $\underline{0.667 \pm 0.052}$ & $0.753 \pm 0.055$             & $0.789 \pm 0.091$ \\
 & \textsc{D2D}             & $0.500 \pm 0.000$             & $0.500 \pm 0.000$             & $0.500 \pm 0.000$             & $0.500 \pm 0.000$             & $0.500 \pm 0.000$             & $0.500 \pm 0.000$ \\
 & \textsc{GraphEraser}     & $0.518 \pm 0.002$             & $0.500 \pm 0.000$             & $0.523 \pm 0.013$             & $0.500 \pm 0.000$             & $0.517 \pm 0.009$             & $0.500 \pm 0.000$ \\
 & \textsc{GraphEditor}     & $0.790 \pm 0.032$             & $0.624 \pm 0.017$             & -                             & -                             & -                             & - \\
 & \textsc{CertUnlearn}     & $0.763 \pm 0.025$             & $0.604 \pm 0.022$             & -                             & -                             & -                             & - \\
 & \textsc{GNNDelete}       & $\underline{0.959 \pm 0.002}$ & $0.964 \pm 0.005$             & $\underline{0.950 \pm 0.002}$ & $0.980 \pm 0.003$             & $\underline{0.924 \pm 0.006}$ & $\underline{0.894 \pm 0.020}$ \\
 & \name (\textit{Ours})    & $\mathbf{0.969 \pm 0.025}$    & $\mathbf{0.752 \pm 0.092}$    & $\mathbf{0.959 \pm 0.013}$    & $\mathbf{0.726 \pm 0.014}$    & $\mathbf{0.939 \pm 0.023}$    & $\mathbf{0.817 \pm 0.019}$ \\
\midrule
\rowcolor[HTML]{EFEFEF}
\multirow{7}{*}{$2.5$}
 & \textsc{Gold}            & $0.965 \pm 0.002$             & $0.777 \pm 0.009$             & $0.955 \pm 0.003$             & $0.739 \pm 0.005$             & $0.934 \pm 0.003$             & $0.858 \pm 0.002$ \\
 & \textsc{GradAscent}      & $0.528 \pm 0.015$             & $0.583 \pm 0.016$             & $0.501 \pm 0.026$             & $\underline{0.576 \pm 0.017}$ & $0.717 \pm 0.022$             & $0.766 \pm 0.019$ \\
 & \textsc{D2D}             & $0.500 \pm 0.000$             & $0.500 \pm 0.000$             & $0.500 \pm 0.000$             & $0.500 \pm 0.000$             & $0.500 \pm 0.000$             & $0.500 \pm 0.000$ \\
 & \textsc{GraphEraser}     & $0.515 \pm 0.002$             & $0.500 \pm 0.000$             & $0.563 \pm 0.013$             & $0.500 \pm 0.000$             & $0.552 \pm 0.009$             & $0.500 \pm 0.000$ \\
 & \textsc{GraphEditor}     & $0.769 \pm 0.040$             & $0.607 \pm 0.017$             & -                             & -                             & -                             & - \\
 & \textsc{CertUnlearn}     & $0.747 \pm 0.033$             & $0.616 \pm 0.019$             & -                             & -                             & -                             & - \\
 & \textsc{GNNDelete}       & $\underline{0.957 \pm 0.003}$ & $\underline{0.892 \pm 0.004}$ & $\underline{0.949 \pm 0.003}$ & $0.905 \pm 0.002$             & $\underline{0.926 \pm 0.007}$ & $\underline{0.898 \pm 0.017}$ \\
 & \name (\textit{Ours})    & $\mathbf{0.965 \pm 0.009}$    & $\mathbf{0.734 \pm 0.055}$    & $\mathbf{0.958 \pm 0.051}$    & $\mathbf{0.718 \pm 0.033}$    & $\mathbf{0.939 \pm 0.026}$    & $\mathbf{0.823 \pm 0.016}$ \\
\midrule
\rowcolor[HTML]{EFEFEF}
\multirow{7}{*}{$5.0$} 
 & \textsc{Gold}            & $0.964 \pm 0.003$             & $0.788 \pm 0.006$             & $0.955 \pm 0.003$             & $0.748 \pm 0.008$             & $0.936 \pm 0.004$             & $0.868 \pm 0.005$ \\
 & \textsc{GradAscent}      & $0.555 \pm 0.099$             & $0.591 \pm 0.065$             & $0.501 \pm 0.023$             & $0.559 \pm 0.024$             & $0.672 \pm 0.032$             & $0.728 \pm 0.022$ \\
 & \textsc{D2D}             & $0.500 \pm 0.000$             & $0.500 \pm 0.000$             & $0.500 \pm 0.000$             & $0.500 \pm 0.000$             & $0.500 \pm 0.000$             & $0.500 \pm 0.000$ \\
 & \textsc{GraphEraser}     & $0.541 \pm 0.002$             & $0.500 \pm 0.000$             & $0.523 \pm 0.013$             & $0.500 \pm 0.000$             & $0.522 \pm 0.009$             & $0.500 \pm 0.000$ \\
 & \textsc{GraphEditor}     & $0.735 \pm 0.037$             & $0.611 \pm 0.018$             & -                             & -                             & -                             & - \\
 & \textsc{CertUnlearn}     & $0.721 \pm 0.033$             & $0.602 \pm 0.013$             & -                             & -                             & -                             & - \\
 & \textsc{GNNDelete}       & $\underline{0.956 \pm 0.004}$ & $\underline{0.859 \pm 0.002}$ & $\underline{0.949 \pm 0.003}$ & $\underline{0.859 \pm 0.005}$ & $\underline{0.924 \pm 0.007}$ & $\underline{0.898 \pm 0.019}$ \\
 & \name (\textit{Ours})    & $\mathbf{0.967 \pm 0.011}$    & $\mathbf{0.753 \pm 0.043}$    & $\mathbf{0.957 \pm 0.028}$    & $\mathbf{0.714 \pm 0.018}$    & $\mathbf{0.940 \pm 0.016}$    & $\mathbf{0.833 \pm 0.015}$ \\
\bottomrule
\end{tabular}
}
\end{table*}

\section{Additional Results}
\label{sec:expdetails}
In addition to the results presented on different GNN architectures namely, GCN~\cite{kipf2016semi}, GAT~\cite{velickovic2017graph} and GIN~\cite{xu2018how}, we also evaluate various datasets, each having different sizes and characteristics. Specifically, we examine PubMed \cite{yang2016revisiting}, Cora \cite{yang2016revisiting} and DBLP \cite{fu2020magnn} using two different edge sampling strategies $\E_f = \E_{f,\mathrm{IN}}$ and $\E_f = \E_{f,\mathrm{OUT}}$ and systematically vary the removal ratios to levels of $0.5\%$, $2.5\%$ and $5.0\%$.

\subsection{PubMed}
The outcomes of the unlearning process involving $0.5\%$, $2.5\%$, and $5.0\%$ edge samples from the 2-hop enclosing sub-graph of $\E_r$ are systematically detailed in Table~\ref{tab:PubMedIN}. The performance of \name surpasses that of existing methodologies in terms of AUC results for the preserved edge set $\E_r$. This underscores \name's \textit{integrity} in retaining knowledge throughout the unlearning procedure across all three GNN architectures. In many instances, the \name's AUC is proximate to that of the \textsc{Gold} model. It surpasses the previous state-of-the-art \textsc{GNNDelete} by up to $4.9\%$. Furthermore, \name effectively minimizes the impact of edges within the forget set $\E_f$ for a significant proportion of cases, highlighting its substantial \textit{consistency}, particularly under higher edge unlearning ratios. While \textsc{GradAscent} and \textsc{GraphEraser} exhibit relatively better outcomes in certain scenarios, their overall knowledge preservation capability is markedly inadequate for meaningful comparison.

The results pertaining to the unlearning of edges beyond the 2-hop enclosing sub-graph of $\E_r$ are systematically outlined in Table~\ref{tab:PubMedOUT}. In this relatively easier scenario, \name consistently outperforms all preceding methodologies across varied unlearning ratios and diverse GNN architectures. This demonstrates that \name is effective for unlearning in various removal ratios. 

\subsection{Cora}
As we shift our focus to the Cora dataset, the findings presented in Tables~\ref{tab:CoraIN} and~\ref{tab:CoraOUT} echo the trends observed in the PubMed dataset. The enhanced performance of \name in terms of both \textit{integrity} and \textit{consistency} reaffirms its effectiveness in different scenarios, unlearning ratios, and GNN architectures. Notably, \name's AUC is notably proximate to that of the \textsc{Gold} model in all instances for the retained set. 

\subsection{DBLP}
The results obtained from the DBLP dataset reinforce the effectiveness of \name in unlearning edges. The findings, detailed in Tables~\ref{tab:DBLPIN} and~\ref{tab:DBLPOUT}, echo the trends observed in the previous datasets. Once again, \name excels in maintaining \textit{integrity} by consistently achieving AUC scores that are remarkably close to those of the \textsc{Gold} model for the retained set.

Overall, the results obtained across different datasets and GNN architectures underline its potential as a valuable tool for unlearning in graph neural networks.

\end{document}